\documentclass{article} 
\usepackage{iclr2026_conference,times}
\usepackage[OT1]{fontenc} 
\usepackage{hyperref}
\usepackage{url}
\usepackage{amsmath,amssymb,bm}
\usepackage{booktabs}
\usepackage{enumitem}
\usepackage{graphicx}
\usepackage{longtable}
\usepackage{xcolor}
\usepackage{eso-pic}
\usepackage{graphicx}
\allowdisplaybreaks
\setlist[itemize]{leftmargin=1.6em}
\setlist[enumerate]{leftmargin=1.6em}
\usepackage{fontspec}
\usepackage{xspace}

\usepackage{fontspec}

\newfontfamily\productfont{Inter}[
    Path           = ./fonts/,
    UprightFont    = Inter-Regular.ttf,
    BoldFont       = Inter-Bold.ttf,
]

\newcommand{\oxygen}{{\productfont\small Oxygen AIIC}\xspace}

\newcommand{\oxygenTitle}{{\productfont Oxygen AIIC}\xspace}
\newcommand{\aiicfullTitle}{{\productfont AI Item Center}}

\title{
\raisebox{-0.265\height}{\includegraphics[width=3.25cm, height=0.65cm]{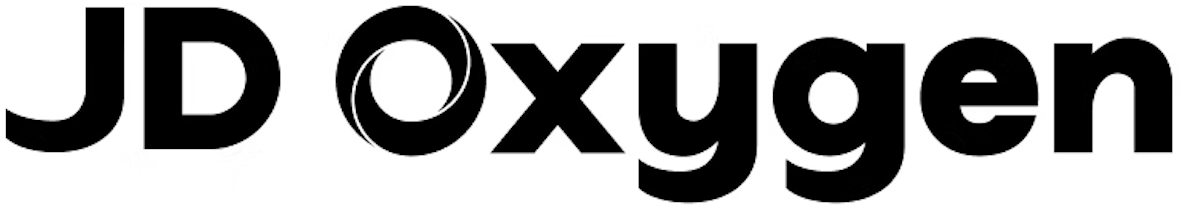}
}
\hspace{-0.25cm}\aiicfullTitle \ (\oxygenTitle) V1: An Industrial-Scale LLM/VLM-Centric Solution for Item Understanding, Management, and Applications
}

\newlength{\authornameskip}
\newlength{\authornameskipfirstline}
\newlength{\authornameskipsecondline}
\newlength{\authornameskipthirdline}
\newlength{\authornameskipforthline}
\newlength{\authornameskipfifthline}
\newlength{\authornameskipsixthline}
\newlength{\authornameskipseventhline}
\newlength{\authornameskipeighthline}
\newcommand{{\ssd}}{\ensuremath{\mathsf{S}^2\mathsf{D}}}

\author{Oxygen AIIC, \hfill Chan Long, \hfill Chao Liu, \hfill Chaofan Chen, \hfill Chaohui Dong, \hfill Chunyuan Guo, \\
\textbf{Danping Liu, \hfill Debin Liu, \hfill Deping Xiang, \hfill Fulai Xu, \hfill Guangyue Liu, \hfill Hao Li, } \\
\textbf{Huichun Hu, \hfill Jian Yang, \hfill Jianan Wang, \hfill Jianbo Zhao, \hfill Jiaoyang Li, \hfill Jiaxing Wang, } \\ 
\textbf{Jinglong Li, \hfill Jinjin Guo, \hfill Jun Fang, \hfill Jun Liu, \hfill Kai Zhou, \hfill Li Wang, \hfill Lili Gao, } \\
\textbf{Liying Chen, \hfill Luning Yang, \hfill Mengdi Zhou, \hfill Pengzhang Liu, \hfill Qi Lv, \hfill Qianyun Wang, } \\
\textbf{Qixia Jiang, \hfill Ruyue Li, \hfill Shimu Liang, \hfill Shuxing Wang, \hfill Sijie Zhang, \hfill Siqi Li, }\\
\textbf{Tianhao Gao, \hfill Wang Ke, \hfill Weihu Huang, \hfill Wencan Lai, \hfill Wenjie Zhang, \hfill Xiaohui Zhang, } \\
\textbf{Xiaojing Dong, \hfill Ya Liu, \hfill Yifeng Zhang, \hfill Yixiang Wang, \hfill Yongtai Zhang, \hfill Yongyi Liao, } \\
\textbf{Zhaoru Chen, \hfill Zhen Chen, \hfill Zhiyong Ma, \hfill Zhiyuan Liu, \hfill Zhongwei Liu, \hfill Ziyan Xing\thanks{Authors are listed in alphabetical order of first name.}} \\
\centerline{\texttt{oxygen-aiic@jd.com}}}

\iclrfinalcopy

\begin{document}

\AddToShipoutPictureFG{%
    \put(
        \LenToUnit{\paperwidth-5.8cm},
        \LenToUnit{\paperheight-1.3cm}
    ){%
        \includegraphics[width=2cm]{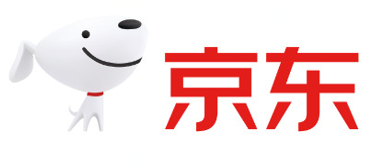}
    }%
}

\maketitle


\begin{abstract}

JD.com\footnote{About JD.com: \url{https://corporate.jd.com/}}, one of the world's largest e-commerce platforms, serves over 700 million active users and millions of merchants, with a catalog of tens of billions of SKUs. At this scale, high-quality, structured item knowledge underpins a better consumer experience, lower management costs, and higher operational efficiency---yet producing and serving it poses three industrial-scale challenges: fast-emerging concepts, high-quality knowledge production for massive SKUs, and diverse downstream requirements. To address these challenges, we present the JD Oxygen AI Item Center (\oxygen), an industrial-scale platform built on LLMs/VLMs for item-knowledge production and service.
\oxygen is built around four core pillars:
(i) ontology engineering driven by efficient human--AI collaboration, which supports the dynamic evolution and agile expansion of an ontology with millions of entries;
(ii) a ``Semantic Search then Discrimination'' ({\ssd}) knowledge-identification architecture that, combined with throughput-improvement strategies, enables scalable, extensible, and high-throughput AI Item Library production for tens of billions of SKUs;
(iii) self-evolving item-understanding LLMs/VLMs that improve in a stable and controllable manner, enabling knowledge production with 94.2\% precision and 82.8\% recall; and
(iv) a unified item tunnel that serves as the data and service hub, delivering item knowledge with tiered freshness.
\oxygen now covers tens of thousands of JD categories and processes hundreds of millions of item updates per day on Huawei Ascend NPUs. It has accumulated hundreds of billions of item-knowledge assets and increased item-information richness to 3.35$\times$ its previous level. Deployed across core business scenarios---including search, recommendation, operations, and category planning---\oxygen has delivered measurable gains at scale. For example, search-traffic coverage reaches 80.4\%, item-information quality issues drop by 37\%, the automated fill rate of core attributes during item listing exceeds 80\%, and intelligent optimization of item creatives increases click-through rate by 9\%.

\end{abstract}

\section{Introduction}

\begin{figure*}[t]
  \centering
    \includegraphics[width=0.9\textwidth,height=0.82\textheight,keepaspectratio]{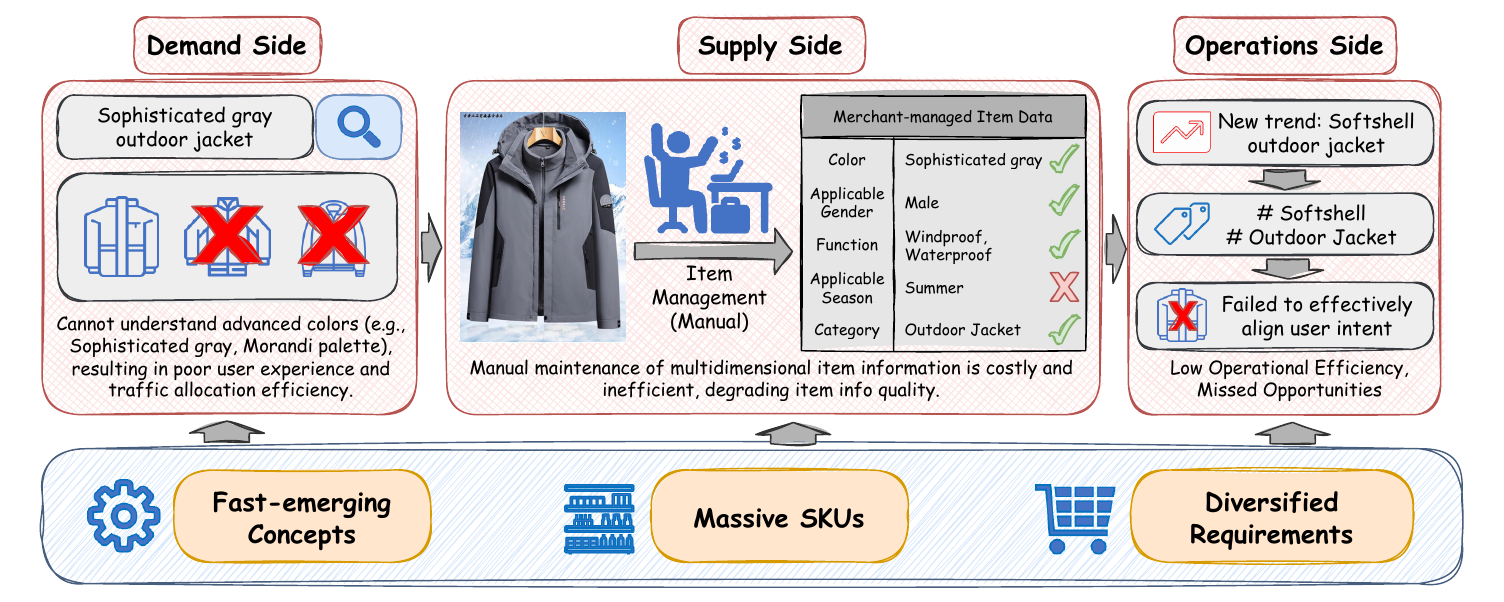}
  \caption{Typical failure cases in traditional item knowledge systems across the demand, supply, and operations sides.}
  \label{fig:badcase}
\end{figure*}

JD.com is a leading e-commerce platform that serves over 700 million active users and millions of merchants, and manages a catalog of tens of billions of SKUs. To deliver on its retail value proposition of
\textit{broader selection, faster delivery, better quality, and lower cost}, 
JD has made ``cost, efficiency, and experience'' its core strategic priorities. 
As e-commerce has grown rapidly, traditional item knowledge systems can no longer support this strategy effectively, giving rise to three industrial-scale bottlenecks across the demand, supply, and operations sides, as illustrated in Figure~\ref{fig:badcase}:

\begin{itemize}
    \item \textbf{Demand side: incomplete item information and semantic gaps.} Incomplete item information, combined with the varied ways users describe their needs---e.g., ``charcoal gray'' vs. ``Morandi palette''---leads to semantic mismatches that degrade the user experience and reduce traffic-allocation efficiency~\citep{nigam2019semantic}.
    \item \textbf{Supply side: costly item management and inefficient traffic acquisition.} Merchants are required to provide and continuously maintain multi-dimensional product information. However, the manual nature of this process makes it costly and inefficient, leading to lower product information quality and limiting merchants' ability to attract traffic.
    \item \textbf{Operations side: fast-changing market trends and growing demand for fine-grained operations.} Frequent trend shifts and increasingly fine-grained operational requirements make trend sensing and item operations more difficult, ultimately constraining the platform's efficiency.
\end{itemize}

To address these bottlenecks, earlier industrial systems adopted traditional NLP techniques and pretrained models, such as BERT-based architectures for named entity recognition (NER)~\citep{luo2020alicoco}. However, these methods remain limited in two major respects. First, owing to their limited model capacity and reliance on task-specific fine-tuning, they struggle to bridge distributional gaps across heterogeneous e-commerce data sources and lack robustness to emerging concepts. Second, they suffer from a ``manual-annotation bottleneck'', in which system accuracy is tightly coupled with costly and unsustainable human labeling. As a result, these methods are difficult to deploy at scale while keeping costs low and quality high.

The rapid progress of LLMs/VLMs offers a way out of the long-standing impasse of high labor cost and weak generalization~\citep{brown2020language,ouyang2022training,radford2021learning,liu2023visual}. Benefiting from extensive world knowledge, strong zero-/few-shot generalization, and reasoning ability, these models enable more accurate, comprehensive, and timely ontology engineering and item-knowledge production.

Academia and industry have advanced intelligent item understanding along four directions: (1) \emph{Domain-specific Foundation Models}, which inject e-commerce knowledge into general-purpose LLMs/VLMs to equip them with retail-domain knowledge~\citep{shi2025llamae,peng2024ecellm,herold2024lilium}; (2) \emph{Automated Ontology Expansion}, which turns the traditional expert-driven paradigm into a semi-automated, human--AI collaboration for dynamic ontology evolution~\citep{shen2020taxoexpan,mao2020octet,errahmadi2023katie}; (3) \emph{Large-scale Attribute Extraction}, which moves item attribute recognition from closed-domain entity extraction to retrieval-augmented generation (RAG) and multimodal understanding~\citep{zheng2018opentag,wang2020aveqa,zhang2025catalograg}; and (4) \emph{Web-scale Item Knowledge Graphs}, which build large, well-aligned retail knowledge networks linking items, attributes, and complex user intents~\citep{luo2020alicoco,luo2021alicoco2,zalmout2021all}.

\begin{figure*}[t]
  \centering
  \includegraphics[width=0.9\textwidth,height=0.82\textheight,keepaspectratio]{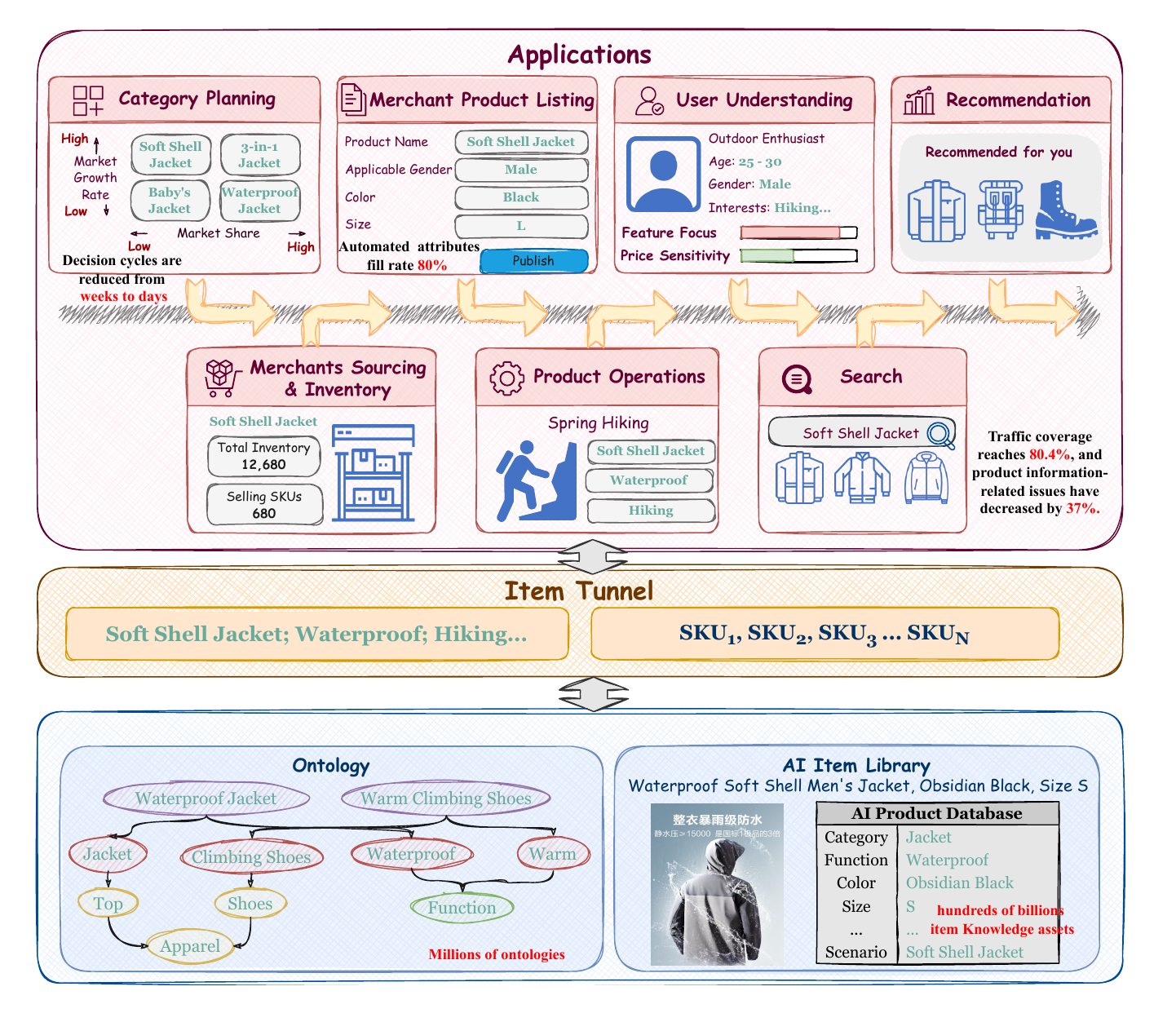}
  \caption{Overview of \oxygen across the item lifecycle. Ontology, and AI Item Library jointly support category planning, merchant workflows, user understanding, search, recommendation, and platform operations.}
  \label{fig:introduction-overview}
\end{figure*}

These efforts confirm the feasibility of large models for intelligent item understanding. However, deploying them at JD, a platform that spans virtually every retail category and manages tens of billions of items, remains challenging. To build a highly available, high-throughput item-knowledge infrastructure, the following three fundamental challenges must be addressed:

\begin{itemize}

\item \textbf{Evolving the ontology to keep pace with heterogeneous sources and fast-emerging concepts.} Item knowledge is multi-source and heterogeneous, scattered across product information (titles, main images, detail pages, etc.), user queries, and public web content. New market segments and concepts emerge constantly, and the required granularity of detail continues to increase. Capturing this knowledge comprehensively and in a timely manner, while expanding the ontology backbone quickly enough to keep pace, is the first challenge for industrial deployment.

\item \textbf{Scalable, high-throughput, low-cost, and high-quality knowledge production at massive item scale.} At the scale of tens of billions of items, the AI Item Library must satisfy several stringent requirements at once: it must scale seamlessly as the ontology evolves, sustain high-throughput production across the full catalog, and keep inference cost and latency within tight budgets, all while keeping knowledge quality consistently high. Directly invoking large models to produce and update item knowledge would incur prohibitive inference cost and unacceptable latency~\citep{dao2022flashattention,kwon2023efficient}, and still fall short on quality. Building an industrial-scale AI Item Library therefore demands a solution that is scalable and extensible by design.

\item \textbf{Efficient support for common and scenario-specific needs.} JD's downstream ecosystem places highly diverse demands on the format and freshness of item knowledge. These scenarios rest on a shared knowledge foundation, yet each carries distinct service requirements: item governance (e.g., information pre-fill and compliance checks) depends on real-time services; search and recommendation need high-throughput nearline features; and category operations require offline post-processing driven by business logic. A single platform must serve all of these highly concurrent, domain-specific demands at once, efficiently building on what the scenarios share while supporting what makes each distinct.
\end{itemize}

To bridge the gap between the potential of LLMs/VLMs and the realities of industrial deployment, we build the JD Oxygen AI Item Center (\oxygen). \oxygen constructs an item ontology with millions of entries and produces high-quality item knowledge at high throughput across tens of thousands of categories and tens of billions of SKUs. It achieves a knowledge-production precision/recall of 94.2\%/82.8\% with a more than 10$\times$ gain in throughput efficiency on Huawei Ascend NPUs, and has accumulated hundreds of billions of clean item-knowledge assets, increasing item-information richness to 3.35$\times$ its previous level. As shown in Figure~\ref{fig:introduction-overview}, \oxygen serves as an item-knowledge hub across the full item lifecycle. In the search scenario, \oxygen covers 80.4\% of traffic and reduces item-information quality issues by 37\%, thereby improving the shopping experience. For category planning, \oxygen shortens decision cycles from weeks to days compared with manual workflows. The automated fill rate of core attributes exceeds 80\%, and optimization of item creatives improves click-through rate by about 9\%.

This paper presents an industrial-scale deployment of LLMs/VLMs for item-knowledge infrastructure. Our main contributions are threefold:
\begin{itemize}
  \item \textbf{An extensible, generalizable, and self-evolving item-understanding LLMs/VLMs framework.} Leveraging the extensive world knowledge and strong reasoning capabilities of large models, we develop item-understanding LLMs/VLMs through incremental learning and model self-evolution, thereby improving knowledge-production quality in a continuous and controllable manner.
  \item \textbf{Knowledge production at the scale of tens of billions of items.} We rapidly enrich the ontology through human--AI collaboration, decouple it from model parameters through the {\ssd} mechanism to accommodate continuous ontology changes and mitigate model hallucinations, and reduce computational cost so that knowledge production at the hundred-billion scale stays efficient.
  \item \textbf{A unified item tunnel and application matrix.} To address diverse requirements across business scenarios, we build an ``item tunnel'' as the shared data and service hub. It maintains data freshness through tiered service levels and, together with the application matrix, supports a wide range of downstream applications, forming a sustainable business ecosystem.
\end{itemize}

\section{Architecture Overview}

\oxygen adopts a modular architecture with high cohesion and low coupling: core capability modules are decoupled and can be iterated independently, improving development efficiency, maintainability, and the platform's ability to evolve in a stable and controlled manner. As shown in Figure~\ref{fig:overall-architecture}, the architecture consists of five tightly coordinated modules:

\begin{figure*}[t]
  \centering
  \includegraphics[width=0.9\textwidth,height=0.78\textheight,keepaspectratio]{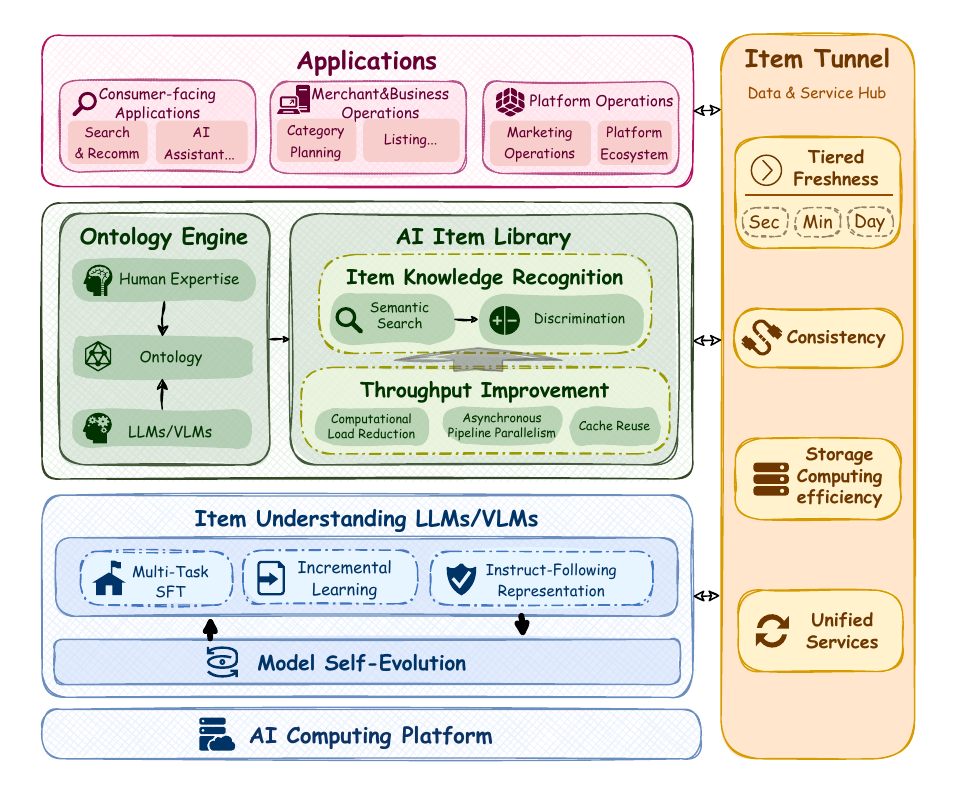}
  \caption{Overall architecture of JD Oxygen AI Item Center. \oxygen integrates ontology engineering, AI Item Library, the item understanding LLMs/VLMs, the item tunnel, and the application matrix into a closed-loop industrial system.}
  \label{fig:overall-architecture}
\end{figure*}

\paragraph{Ontology Engineering}
The ontology is the knowledge foundation of \oxygen and determines the upper bound of item-knowledge quality and application potential. Through efficient human--AI collaboration, \oxygen combines more than 20 years of JD expert knowledge with the world knowledge and reasoning capabilities of LLMs/VLMs to produce a high-quality, comprehensive, and timely ontology. Experts focus on distilling industry knowledge, while algorithms learn from it to scale ontology construction and drive continuous evolution.

\paragraph{AI Item Library}
The AI Item Library maps items to the ontology and serves as the source of item knowledge for downstream applications. Given a continuously evolving ontology and tens of billions of items, we achieve scalable, high-throughput production by constructing a jointly optimized model-data-engineering pipeline. 
Decoupling the ontology from the model parameters reduces hallucinations and improves generalization, while computational load reduction, cache reuse, and asynchronous pipeline parallelism ensure efficient production at scale.

\paragraph{Item-Understanding LLMs/VLMs}
The item-understanding LLMs/VLMs support both ontology construction and AI Item Library production, serving as the foundation for continuous improvement in \oxygen's data quality. We integrate the algorithmic capabilities required by \oxygen into a highly generalizable and scalable foundation model. Through incremental learning and model self-evolution, the system fills targeted knowledge gaps and mitigates catastrophic forgetting, enabling model capabilities to evolve in a stable and controlled manner.

\paragraph{Item Tunnel}
The item tunnel is the central hub between \oxygen and business applications, providing a unified service layer. To meet downstream requirements for different levels of freshness and throughput, it supports daily-, minute-, and second-level production and distribution pipelines while preserving data consistency. It enables downstream applications to consume \oxygen capabilities efficiently.

\paragraph{Applications}
The application matrix is where \oxygen delivers its value. It turns the item-knowledge assets and model capabilities exposed by the tunnel into standardized services and deploys them at scale across business formats, scenarios, and the end-to-end item lifecycle. It bridges technology and business, serving as core infrastructure for the platform's AI-driven e-commerce transformation and sustainable growth.

In summary, the five modules of \oxygen are tightly integrated rather than operating in isolation. Together, they form an end-to-end loop spanning large-scale ontology construction, massive knowledge production, centralized asset management, tiered access, and cross-domain feedback. This loop preserves stable, high-quality item knowledge while allowing the system to evolve continuously. The following sections introduce each module in detail.

\section{Ontology Engineering}
Ontology engineering aims to build a high-quality, comprehensive, and timely item-knowledge foundation~\citep{luo2020alicoco, luo2021alicoco2, yu2024cosmo, yu2023folkscope, dong2020autoknow, huang2025attributeforge}. To achieve this goal, it is essential to effectively combine the domain expertise accumulated by JD over more than two decades with the large-scale concept-mining capabilities of LLMs/VLMs. However, several practical challenges arise in this process:
\begin{itemize}
  \item \textbf{Continuously emerging concepts:} The volume of new concepts emerging daily makes fully manual ontology construction increasingly inadequate in both timeliness and coverage.
  \item \textbf{High semantic redundancy:} Knowledge extracted from heterogeneous data sources often contains numerous synonymous or highly overlapping concepts, which can quickly inflate the ontology and reduce its consistency.
  \item \textbf{Achieving scale and quality at once:} Ontology construction must simultaneously scale to broad coverage and maintain high, controllable quality. Expert-driven construction ensures quality but cannot scale, whereas large-scale automated generation by LLMs/VLMs scales but, without sufficient oversight, introduces hallucinations and quality issues.
\end{itemize}

To address these challenges, we adopt a human--AI collaborative framework in which experts define ontology standards and perform final quality audits, while LLMs conduct large-scale ontology discovery, expansion, and refinement under expert guidance. This framework enables the ontology to evolve continuously while maintaining both high quality and broad coverage.

\subsection{Method Overview}

\begin{figure*}[t]
  \centering
  \includegraphics[width=0.9\textwidth,height=0.78\textheight,keepaspectratio]{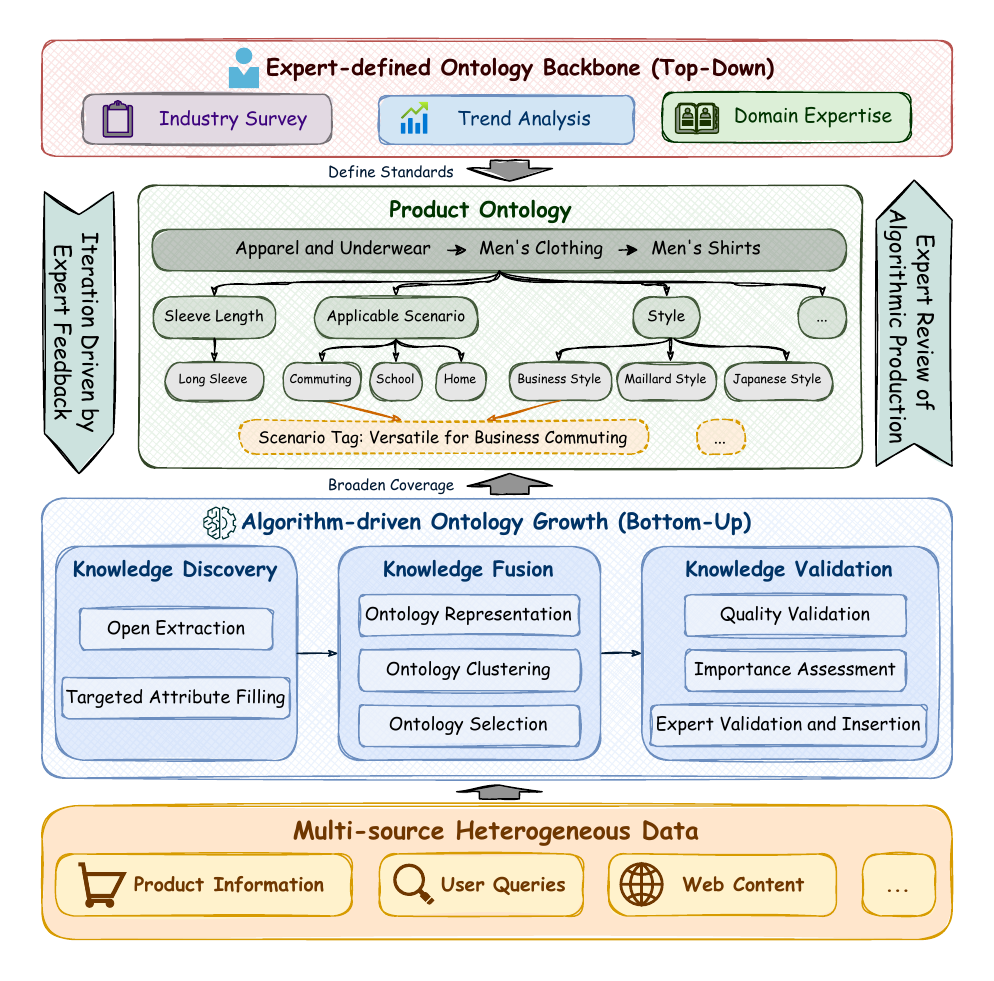}
  \caption{Human--AI collaborative ontology engineering. Human experts establish the fundamental ontology backbone, while an automated pipeline dynamically discovers, fuses, and validates emerging concepts from multi-source heterogeneous data. }
  \label{fig:ontology-engineering}
\end{figure*}

As shown in Figure~\ref{fig:ontology-engineering}, the ontology is organized around downstream business requirements. Experts define four core element types: (i) \emph{Category}, the item taxonomy (e.g., apparel and underwear $\rightarrow$ men's clothing $\rightarrow$ men's shirts); (ii) \emph{Attribute Key}, an item feature dimension (e.g., sleeve length for men's shirts); (iii) \emph{Attribute Value}, a specific instantiation of an attribute (e.g., long sleeve for sleeve length); and (iv) \emph{Scenario Tag}, a higher-level composite concept that captures a consumption context (e.g., World Cup Watch Party Bundle). Categories, attribute keys and attribute values constitute the backbone of the ontology, carrying atomic item knowledge that supports the platform's core business operations. Scenario tags provide an additional semantic layer by aggregating atomic knowledge into higher-level concepts, capturing multi-dimensional semantic relationships and enabling rapid support for downstream scenario-based demands.

Our human–AI collaborative framework operates in an expert-guided, AI-driven loop. Initially, experts establish the ontology backbone, which supplies structural prior knowledge and serves as a semantic anchor. Based on this foundation, the automated pipeline implements a three-stage workflow: knowledge discovery, fusion, and validation. This pipeline continuously harvests emerging high-frequency concepts from heterogeneous data sources and integrates them into the ontology, ensuring scalable and dynamic infrastructure evolution.

\subsection{Ontology Construction}  \label{sec:3_2}

\subsubsection{Expert-defined ontology backbone (top-down)}
The category knowledge accumulated by JD's domain experts provides a strong foundation for identifying the item attributes that influence purchase decisions and capture business-relevant distinctions. To transform this expertise into structured priors that can guide algorithmic discovery, we establish a standardized expert workflow that defines the backbone of the ontology. Specifically, experts curate the major category hierarchy together with its core attribute sets, representative attribute values, and characteristic scenario tags.

At this stage, experts focus only on high-value and representative knowledge rather than pursuing exhaustive coverage. This top-down design effectively constrains the space of knowledge generation and establishes clear semantic boundaries for the algorithms' automated, large-scale knowledge mining~\citep{lippolis2024ontogenia, saeedizade2024navigating, lippolis2025ontology, babaei2023llms4ol, mateiu2023ontology, fathallah2024neon, fathallah2024llms4life, sun2025lkd}.

\subsubsection{Algorithm-driven ontology growth (bottom-up)} \label{sec:3_2_2}

Building upon the expert-defined ontology backbone and continuously incorporating signals from user behavior and industry trends, the algorithms expand the ontology at scale through a bottom-up ``discovery--fusion--validation'' pipeline. Taking attribute-value expansion as an example, the discovery stage identifies emerging concepts from heterogeneous data sources; the fusion stage consolidates synonymous and semantically related concepts into normalized candidates; and the validation stage evaluates each candidate for both quality and business importance. Guided by continuous expert feedback, validated concepts are incorporated into the ontology in a controlled and scalable manner, enabling its sustained evolution while preserving high quality~\citep{edge2024local, tiwari2025ontorag, zhang2024extract, ye2023schema, bai2025autoschemakg}.

\paragraph{(1) Knowledge discovery.} 

Knowledge discovery aims to identify latent concepts from large-scale heterogeneous data. Although LLMs/VLMs can assist with this task, general-purpose models are not explicitly aligned with the e-commerce ontology and often fail to capture domain-specific concepts, industry standards, and emerging terminology. To address this limitation, we train a dedicated knowledge-discovery model for the e-commerce domain.

\textbf{Training data construction.} To support latent-concept discovery, we organize the training data into a unified triple format $\langle x, k, v \rangle$, where $x$ denotes product information, user queries, or external web content; $k$ denotes an attribute key; and $v$ denotes the corresponding attribute value. We construct the training data using state-of-the-art LLMs/VLMs with hundreds of billions of parameters (strong reasoning and generalization capabilities), via two strategies: open information extraction (OpenIE) \citep{wang2023instructuie, gui2023instructie, han2023empirical} and targeted attribute filling \citep{lu2022unified}.

(i) \textbf{OpenIE:} This method is suitable for mining large-scale corpora. The model discovers potential attribute keys $k_i$ and their corresponding attribute value sets $V_{k_i}$ from $x$:
\[
f_{\text{OpenIE}}(x) \rightarrow \{(k_1, V_{k_1}), (k_2, V_{k_2}), \dots, (k_n, V_{k_n})\}.
\]
For each extracted pair $(k_i,V_{k_i})$, we combine it with the original input $x$ to form triple samples:
\[
\mathcal{T}_{\text{OpenIE}}=\{\langle x,k_i,v_{ij}\rangle \mid v_{ij} \in V_{k_i}\}.
\]
For example, given the item title ``summer ice-silk sun-protective cardigan'', the model extracts \{``material'': ``ice silk'', ``applicable season'': ``summer'', ``function'': ``sun protection''\}.

(ii) \textbf{Targeted attribute filling:} This method is implemented as an attribute-specific NER task and is suitable for completing attribute values for expert-defined or core attributes. Given an attribute $k$ and its business definition $d_k$, the model is required to extract $V_k$ for that attribute from $x$:
\[
f_{\text{NER}}(x, k, d_k) \rightarrow \{k: V_k\}.
\]
Each $v_j$ in $V_k$ is similarly converted into a triple sample:
\[
\mathcal{T}_{\text{NER}}=\{\langle x,k,v_j\rangle \mid v_j\in V_k\}.
\]
For example, given the input ``men's business commuting non-iron three-quarter-sleeve white shirt'' and the key ``sleeve length'' with the description ``including long sleeve, short sleeve, etc.'', the model extracts sleeve length: [``three-quarter-sleeve''].

\textbf{Discovery model training.} We adopt a unified supervised fine-tuning (SFT) framework and formulate the two tasks above as follows:
\[
f_{\text{ext\_kv}}(x, I_{\text{OpenIE}}) \rightarrow \{k_i: [v_{i1}, v_{i2}, \dots]\},
\]
\[
f_{\text{ext\_v}}(x, k, d_k, I_{\text{NER}}) \rightarrow\{v_1,v_2,\dots,v_m\}.
\]
Both $I_{\text{OpenIE}}$ and $I_{\text{NER}}$ denote task instructions that explicitly constrain the model's extraction objective. After training, the model achieves 91\% precision and 79\% recall on the candidate knowledge extraction task.

Unless otherwise stated, all experiments use an 8B-parameter base model. We observe that after SFT, different base models exhibit comparable performance, with variations of less than 2\% across evaluation metrics. Therefore, the remainder of the paper focuses on the framework rather than base-model selection.

Applying the trained discovery model to multi-source corpora, including item information, user queries, and external web content, yields approximately 4.5 million latent attribute values. Each value is represented as a standardized knowledge unit $\langle c,x,k,d_k,v\rangle$, where $c$ denotes the category.

\paragraph{(2) Knowledge fusion.} 

Knowledge discovery improves ontology coverage but also introduces a large number of heterogeneous synonymous expressions, resulting in redundancy. To address this issue, we adopt a three-stage knowledge fusion strategy consisting of representation, clustering, and selection. 
\begin{itemize}
    \item Representation stage: latent concepts are encoded into vector representations.
    \item Clustering stage: semantically identical entities are grouped together.
    \item Selection stage: a standard ontology concept is extracted from each cluster.
\end{itemize}

Because general-purpose representation models lack sufficient understanding of e-commerce semantics, we train a domain-adapted encoder. The encoder captures contextual semantics in e-commerce scenarios, providing high-fidelity and effective representations of latent concepts. This section consists of three parts: training data construction, representation model training, and ontology fusion.

\textbf{Training data construction.} Given the input $\langle c, x, k, d_k, v \rangle$ above, the instruction for the encoder is defined as $I_{\text{fuse}}(c,k,d_k)$. This instruction explicitly injects category $c$, attribute key $k$, and attribute description $d_k$ into the encoder as context, so that attribute value $v$ is encoded as:
\[
f_{\text{enc}}\bigl(v,\, I_{\text{fuse}}(c,k,d_k)\bigr) \rightarrow \mathbf{e}
\]
Training sample construction focuses on two questions: which ontology entries should be pulled closer as positives, and which should be pushed apart as negatives. For an attribute value $v$, the LLM generates $N$ synonymous rewrites as the positive set $V^+=\{v^+_1,\dots,v^+_{|V^+|}\}$. For the current $\langle c, k\rangle$, a general encoder retrieves expressions similar to $v$, and an LLM then judges synonymy among the retrieved results. Synonymous pairs are removed, while the near-neighbor but non-synonymous expressions are retained as $V^-=\{v^-_1,\dots,v^-_{|V^-|}\}$.

Each training sample consists of an anchor $v$, a positive set $V^+$, and a negative set $V^-$. All samples share the same category--attribute context $\langle c, k, d_k\rangle$, forcing the encoder to learn fine-grained semantic distinctions among attribute values. To improve representation quality for infrequent concepts, long-tail attribute values are additionally oversampled during training.

\textbf{Representation model training.} The representation model adopts a bi-encoder architecture with shared parameters across the two towers and an 8B-parameter LLM as the backbone. Given an input sequence, we append a special $\langle\mathrm{eos}\rangle$ token to its end and use the final-layer hidden state at this position as the sequence representation. The resulting embedding is $L_2$-normalized, yielding the final sentence vector $\mathbf{e}$. The training objective is the InfoNCE loss \citep{oord2018representation}:
\[
\mathcal{L}_{\text{InfoNCE}} = -\frac{1}{|V^+|}\sum_{i=1}^{|V^+|} \log \frac{\exp\bigl(\mathrm{sim}(\mathbf{e},\, \mathbf{e}^{+}_{i}) / \tau\bigr)}{\exp\bigl(\mathrm{sim}(\mathbf{e},\, \mathbf{e}^{+}_{i}) / \tau\bigr) + \sum_{j=1}^{|V^-|} \exp\bigl(\mathrm{sim}(\mathbf{e},\, \mathbf{e}^{-}_{j}) / \tau\bigr)}.
\]
Here, $\mathbf{e}$, $\mathbf{e}^{+}_{i}$, and $\mathbf{e}^{-}_{j}$ are the vectors obtained by encoding the anchor $v$, the $i$-th positive sample $v^+_i$, and the $j$-th negative sample $v^-_j$ through $f_{\text{enc}}$, respectively; $\mathrm{sim}(\cdot,\cdot)$ denotes cosine similarity, and $\tau$ denotes the temperature coefficient. The loss is averaged over all $|V^+|$ positive samples.

On the ontology-similarity test set, our encoder raises the Spearman correlation coefficient from 0.62 (general-purpose encoder) to 0.86.

\textbf{Ontology fusion.} 
Based on the representation model above, all candidate ontology units $\langle c, k, v\rangle$ are processed as follows:
\begin{itemize}
  \item Representation: each attribute value is encoded into a vector within its corresponding category--attribute context $\langle c,k\rangle$.
  \item Clustering: vectors belonging to the same $\langle c,k\rangle$ subspace are grouped using hierarchical clustering \citep{murtagh2014ward} to identify semantically equivalent concepts.
  \item Selection: within each cluster, candidate terms are ranked based on their occurrence frequency and LLM-driven semantic evaluation. The highest-ranked term, $v^*$, is designated as the canonical ontology entry, while the remaining terms form its synonym set, $V_{\text{syn}}$.
\end{itemize}

The fusion stage produces standardized ontology candidates of the form $\langle c, x, k, d_k, v^*, V_{\text{syn}}\rangle$, which are subsequently passed to the validation stage. Applying the fusion pipeline reduces the number of discovered concepts from 4.5 million to 2.1 million candidate ontology concepts, consolidating approximately 2.4 million redundant concepts into their corresponding synonym sets $V_{\text{syn}}$.

\paragraph{(3) Knowledge validation.}

Knowledge validation aims to ensure the quality of ontology candidates generated through large-scale discovery and fusion. While task-specific models achieve strong performance within their training distribution, they often suffer from limited generalization when confronted with emerging concepts and distribution shifts \citep{huang2025empirical, wataoka2024self}. In contrast, general LLMs exhibit stronger open-domain reasoning capabilities and broader semantic coverage, enabling more reliable assessment of candidate ontology concepts.

To combine these complementary strengths, we propose a Multi-LLM Collaborative Verification Framework that integrates multiple LLM validators with expert knowledge. The framework evaluates candidate ontology concepts through a sequence of quality and business-importance assessments, ensuring that only high-quality, important concepts are incorporated into the ontology.

\textbf{Knowledge quality validation.} To strictly control ontology quality, the validation process follows a sequential procedure: a candidate ontology entry must pass plausibility and duplication checks in order. If any step returns ``Reject'', the process stops and the candidate is blocked.

\begin{itemize}
    \item \textbf{Plausibility validation:} the system checks whether the candidate makes common sense and satisfies category--attribute constraints. For example, ``spicy-flavored computer'' violates common sense and is rejected outright; ``operating system is Android'' under ``laptop computer'' violates category constraints and is also rejected.
    \item \textbf{Duplication validation:} the system incrementally compares the candidate with existing ontology entries to prevent duplicate insertion. For example, if ``CPU model'' already exists, ``central processing unit model'' is rejected.
\end{itemize}

To reduce variation and one-off misjudgments from any single model, each validation stage introduces a multi-model majority-voting mechanism \citep{schoenegger2024wisdom}: $M$ general LLMs are called simultaneously, and each model independently outputs $y_i \in \{\text{Pass}, \text{Reject}, \text{Unclear}\}$. The final decision follows majority voting:
\[
y =
\begin{cases}
\text{Pass}, & \text{if } \sum_{i=1}^{M} \mathbb{I}(y_i = \text{Pass}) > \dfrac{M}{2},\\[4pt]
\text{Reject}, & \text{otherwise.}
\end{cases}
\]
where $\mathbb{I}(\cdot)$ is the indicator function and $y_i$ is the judgment of the $i$-th model. Only when a majority of models output ``Pass'' does the candidate enter the next stage; otherwise, it is immediately blocked and sent to an exception or manual-review pool.

\textbf{Knowledge importance assessment.} After passing quality validation, candidate knowledge is further assessed for its business importance and future growth potential to determine its insertion priority and downstream processing strategy. Inspired by the Boston Matrix, we characterize each candidate along two dimensions: \emph{scale} and \emph{trend}.
\begin{itemize}
  \item \textbf{Scale:} measures the current prevalence of a candidate concept within the e-commerce ecosystem, including item coverage and query frequency. Concepts with high scale primarily help broaden ontology coverage.
  \item \textbf{Trend:} measures the future growth potential of a candidate concept, including short-term frequency growth and changes in search popularity. Concepts with high trend help the ontology track emerging hot topics.
\end{itemize}
Using these two dimensions, the system classifies candidate knowledge into four types: high trend $\times$ high scale (star concepts), high trend $\times$ low scale (emerging-trend concepts), low trend $\times$ high scale (stable general concepts), and low trend $\times$ low scale (low-value concepts). By default, low-value concepts are not inserted automatically.

\textbf{Expert validation and insertion.} After completing the importance assessment, the system performs tiered ontology insertion based on the evaluation results. Candidate knowledge that receives a unanimous \textit{Pass} during the quality validation stage is automatically inserted into the ontology. Candidate knowledge with \textit{Unclear} judgments or substantial disagreement across models is forwarded to domain experts for final confirmation and boundary determination.

In summary, the knowledge validation module focuses not only on whether candidate knowledge is correct, but also on whether it is important. By keeping the ontology accurate, comprehensive, and responsive to emerging hot topics at once, the framework supports continuous and controllable ontology evolution.

\subsection{Results}
Under the human--AI collaborative framework, the system has built a million-scale, high-quality ontology encompassing attributes, product terms, brands, and other entities. Compared with the previous generation, the average number of characterization dimensions per item has increased to 1.44$\times$ the previous level, and the ontology scale has expanded by 64.5\%, significantly enhancing product information richness. The resulting ontology covers 80.4\% of JD's user traffic.

\section{AI Item Library}
The ontology provides the system with a standardized knowledge foundation, while the core mission of the AI Item Library is to achieve a scalable and extensible semantic mapping from large-scale unstructured item information to the ontology. Essentially, it identifies ontology elements from items with high precision, thereby building a stable item-to-ontology association chain.

The data carried by this association chain exhibits typical industrial-scale characteristics, covering tens of billions of SKUs, tens of thousands of categories, and millions of dynamically evolving ontology entries. It must handle a daily stream of hundreds of millions of item-information changes while supporting JD's core application scenarios with high freshness. This leads to two core technical challenges:
\begin{itemize}
  \item \textbf{Scalability under dynamic ontology.} In conventional end-to-end models, ontology knowledge is tightly coupled with model parameters, making adaptation to ontology updates costly and often leading to degraded out-of-distribution (OOD) performance. Alternatively, adapting to ontology evolution through frequent fine-tuning or retraining incurs prohibitive computational and time costs.

  \item \textbf{Throughput bottleneck under massive data.} Exhaustively evaluating all attributes for tens of billions of SKUs is highly inefficient. Large groups of homogeneous SKU variants incur redundant processing, while most computations on sparse long-tail attributes yield no useful signals.
\end{itemize}

\subsection{Method Overview}

Existing methods for item knowledge recognition \citep{chen2023does, shinzato2022simple,yang2022mave,yan2021adatag} still face significant structural limitations in real-world industrial scenarios. Extraction-and-mapping methods first extract non-standardized values using entity recognition \citep{zheng2018opentag,xu2019scaling,yan2021adatag} or question-answering models \citep{wang2020learning,yang2023mixpave}, and then map them to standard values. Nevertheless, the ontology knowledge is often implicitly encoded in model parameters, making such methods difficult to adapt to ontology updates and leading to degraded generalization. Classification-based methods \citep{chen2022extreme} formulate attribute value recognition as a multi-label classification problem over a closed label space. Consequently, they fail to accommodate newly added open-domain attributes, and each ontology update may require model retraining, making them poorly suited to the dynamically evolving attribute systems of e-commerce platforms. Generative models \citep{sabeh2024empirical,nikolakopoulos2023sage,shinzato2023unified} exhibit stronger generalization capabilities and can partially mitigate OOD issues, but their outputs are difficult to constrain and are susceptible to model hallucination during direct attribute value generation. Finally, similarity-based retrieval methods \citep{su2025taclr} select candidate values via semantic vector matching, but their effectiveness is highly sensitive to threshold settings, resulting in diminished robustness in production environments.

Therefore, scalability and high throughput must be treated as joint design objectives to enable effective implementation and long-term stability in industrial-scale business scenarios. In light of these considerations, we propose a collaborative optimization system spanning models, data, and engineering.

\begin{figure*}[t]
  \centering
  \includegraphics[width=0.9\textwidth,height=0.70\textheight,keepaspectratio]{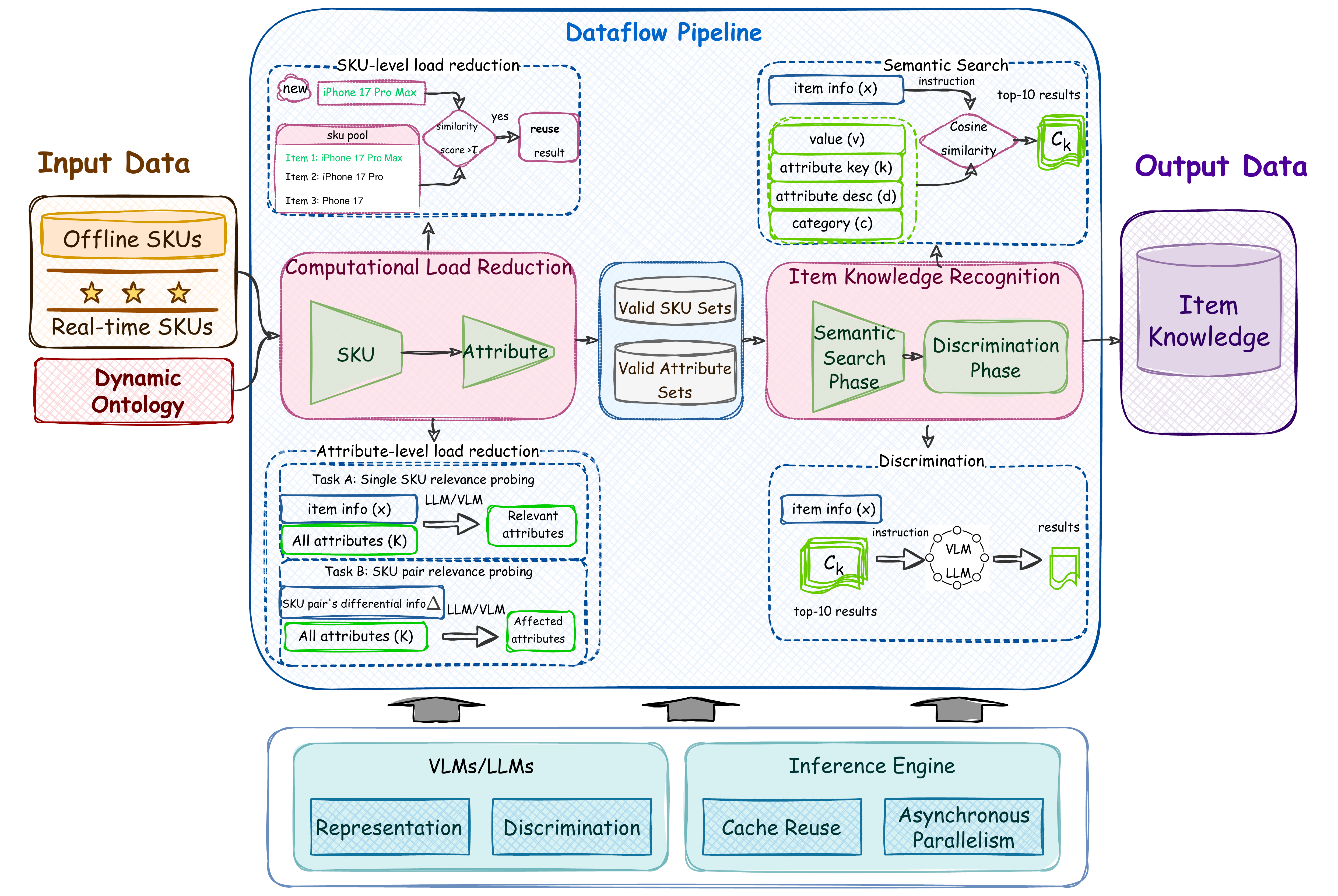}
  \caption{Production architecture of the AI Item Library. Taking item data and a dynamically evolving ontology as input, the pipeline first mitigates computational redundancy across the SKU and attribute dimensions, and then performs precise item-to-ontology recognition through a two-stage ``Semantic Search then Discrimination'' ({\ssd}) engine, powered by the item understanding LLMs/VLMs.}
  \label{fig:ai-item-knowledge-base-production}
\end{figure*}

As shown in Figure~\ref{fig:ai-item-knowledge-base-production}, we propose an industrial-scale ``Semantic Search then Discrimination'' ({\ssd}) architecture \citep{zou2025multi} that decouples the ontology from model parameters. In the semantic search stage, the dynamically evolving ontology is externalized as a separate ontology knowledge base, enabling continuous ontology updates without model retraining. Semantic encoders retrieve ontology entries relevant to the given item. In the discrimination stage, the model only determines whether the item matches the retrieved ontology entries. This formulation substantially reduces task complexity, mitigates model hallucination, and enhances generalization to ontology evolution.

To further improve efficiency, we apply computational load reduction across both the SKU and attribute dimensions. In the SKU dimension, similarity-based deduplication eliminates redundant computation over homogeneous item variants. In the attribute dimension, the conventional ``full-attribute scanning'' paradigm is reformulated as ``highly relevant attribute probing'', where the {\ssd} process is initiated only for attributes deemed relevant. In addition, cache reuse and asynchronous pipeline parallelism are employed to minimize redundant computation, optimize NPU utilization, and improve overall system throughput.

\subsection{Item Knowledge Recognition}  \label{sec:4_2}

The {\ssd} architecture formulates item knowledge recognition as a two-stage process. The detailed implementation is described below.

\subsubsection{Semantic Search Stage} \label{sec:4_2_1}

The primary objective of this stage is to model the matching relationships between items and ontology entries. Given item information, the system retrieves the Top-$K$ most relevant ontology entries from the dynamically evolving ontology as candidates. As in Section~\ref{sec:3_2_2}, this stage injects domain knowledge into the representation model, aligning items and ontology entries within a unified semantic space.

\paragraph{(1) Representation model training.} ~\\

\textbf{Data construction.}
In contrast to Section~\ref{sec:3_2_2}, this stage adjusts both the encoding inputs and the positive/negative sample construction method.

For item encoding, the category $c$ serves as contextual information and is injected through the instruction template $I_{\text{sku}}(c)$. Given an item $x$, its representation is derived as:
\[
f_{\text{enc}}\bigl(x,\, I_{\text{sku}}(c)\bigr) \rightarrow \mathbf{e}_{\text{sku}}.
\]

For ontology encoding, the category $c$, attribute key $k$, and attribute description $d_k$ serve as contextual information and are injected through the instruction template $I_{\text{val}}(c,k,d_k)$. Given an attribute value $v$, its representation is derived as:
\[
f_{\text{enc}}\bigl(v,\, I_{\text{val}}(c,k,d_k)\bigr) \rightarrow \mathbf{e}_{\text{val}}.
\]

Both input types share the same encoder parameters $f_{\text{enc}}$ and are differentiated solely by their respective instruction templates.

\textbf{Positive and negative sample construction.}
Positive samples $V_k^+$ are derived from high-confidence results generated during knowledge discovery. Negative samples $V_k^-$ are drawn from other attribute values under the same category $c$ and attribute key $k$. These candidate values are further screened by a large language model, and only those that are not semantically grounded in the item $x$ are designated as negative samples. This strategy prioritizes standard values within the same category and attribute, which are the hardest to distinguish and therefore provide highly informative contrastive signals.

\textbf{Model training.}
The model architecture, loss function, and training strategies are consistent with those described in Section~\ref{sec:3_2_2}. Furthermore, we perform multi-task joint training to consolidate interrelated capabilities into a unified model, using instruction templates $I_{\text{fuse}}$, $I_{\text{sku}}$, and $I_{\text{val}}$ to distinguish the different tasks.

This design aligns items and ontology entries within a unified semantic space. Moreover, because the large-scale representation model generalizes robustly, newly added ontology entries can be directly encoded by the same encoder and integrated into retrieval without model retraining.

\paragraph{(2) Knowledge retrieval.} ~\\

\textbf{Retrieval with the representation model.}
For offline retrieval, the system maintains two vector indexes generated by the same encoder. The first is an attribute-value index, which encodes attribute values under each category and attribute key. The second is an item index, which encodes comprehensive item information. Both indexes store vector representations along with their corresponding metadata and support incremental updates. Consequently, newly added ontology entries can be rapidly integrated without triggering model retraining.

During semantic search, for each target category and attribute key, the system calculates the cosine similarity between the item representation and the attribute-value representations. Then, it retrieves the Top-$K$ candidate attribute values:
\[
C_k = \{v_1, v_2, \dots, v_K\},
\]
where $K=10$ is empirically chosen to achieve strong recall while keeping downstream discrimination cost low. Thus, the initial candidate space is reduced to a compact set of Top-$K$ candidates, significantly alleviating the computational burden of the subsequent discrimination stage.

\subsubsection{Discrimination Stage}

Using the Top-$K$ candidate set from the semantic search stage, the discrimination stage performs fine-grained matching between the item and its retrieved candidates. Although general LLMs/VLMs contain hundreds of billions of parameters, their lack of domain-specific knowledge prevents them from achieving high precision, typically keeping it below 80\%. Such models often cannot meet throughput and recognition-accuracy requirements at the same time, motivating the need for a specialized discrimination model. This section comprises two primary phases: discrimination model training and knowledge recognition.

\paragraph{(1) Discrimination model training.} ~\\

\textbf{Training data construction.}
Based on the extraction training data constructed in Section~\ref{sec:3_2_2}, given item information $x$ and a target attribute key $k$, the input to the discrimination model is formalized as $\langle x, k, C_k\rangle$, where $C_k$ represents the candidate attribute value set under $k$. The model is tasked with identifying which values in $C_k$ are semantically consistent with the item and outputting the corresponding matching subset $C_k^* \subseteq C_k$.

Training data construction consists of two steps: constructing the candidate set $C_k$ and deriving the supervised label set $C_k^*$. The candidate set comprises potential positive samples and same-attribute negative samples:
\[
C_k = V_k^+ \cup V_k^-,
\]
where $V_k^+$ denotes the set of potential positives and $V_k^-$ denotes the set of negatives. The construction of $V_k^+$ and $V_k^-$ aligns with the strategy employed for representation model training within the semantic search stage, as described in Section~\ref{sec:4_2_1}.

After obtaining $C_k$, we further generate the supervised output $C_k^*$ to guide model training. Since the candidate values are highly relevant to the target attribute key, standard off-the-shelf LLMs are prone to generating false positives or false negatives when encountering closely related values, implicit item descriptions, or ambiguous boundary samples. Therefore, we first fine-tune an industrial-scale foundation model on a small set of high-quality data to derive a reference model $M_T$. This reference model offers more stable attribute value recognition and stronger boundary discrimination.

To construct large-scale supervision data, we use the reference model $M_T$ to generate pseudo labels, distilling its discrimination capability into the training corpus:
\[
M_T(x, k, C_k) \rightarrow C_k^*.
\]

To further improve robustness in production scenarios, we implement multi-granularity data augmentation during training, including candidate-set augmentation and sample-distribution calibration. Candidate-set augmentation randomly shuffles the order of candidate values, mitigating the model's reliance on spurious correlations arising from candidate sequencing, attribute combinations, or template positions. Sample-distribution calibration controls the ratio of positive to negative samples, preventing the model from overfitting to distributions dominated by negative candidates and thereby avoiding overly conservative rejection behavior.

\textbf{Model training.}
The discrimination model is trained via supervised fine-tuning. Let $I_{\text{dis}}(x,k)$ represent the discrimination instruction formulated from item information $x$ and the target attribute key $k$. Conditioned on the candidate set $C_k$, the model is optimized to generate the supervised label set $C_k^*$:
\[
f_{\text{dis}}\bigl(C_k, I_{\text{dis}}(x,k)\bigr) \rightarrow C_k^*.
\]

Through this training paradigm, the model learns to perform set-based filtering within a constrained candidate space. This design minimizes the risk of model hallucination and enhances output controllability, as the model is restricted to selecting values from the ontology-constrained candidate set instead of freely generating unconstrained attribute values. Furthermore, the expertise of the large-scale reference model is transferred to a more efficient 8B-parameter model, substantially reducing inference cost in production scenarios.

\paragraph{(2) Knowledge recognition.} ~\\

During online knowledge recognition, given an item $x_i$ and an attribute key $k$, the recall module initially returns the Top-$K$ candidate attribute value set $C_k$. The discrimination model then uses the formulation above to select the subset $C_k^*$ that is semantically consistent with the item. If no candidate value matches the item information, the model outputs an empty set, indicating that the item has no compatible attribute value for the specified $k$.

Finally, the predicted result is converted into structured item knowledge:
\[
\mathcal{R}_{i,k}
=
\{\langle x_i, k, v\rangle \mid v \in C_k^*\}.
\]

Thus, the discrimination model performs precise knowledge recognition within the recalled candidate space. The output is strictly constrained by the item ontology, which enhances system reliability and mitigates model hallucination. Furthermore, newly added ontology entries can be readily incorporated into the online recognition workflow as soon as they are encoded into the ontology index.

\subsubsection{Results}  \label{sec:4_2_3}

To assess the efficacy of {\ssd}, we conducted random sampling of items across diverse categories to evaluate their knowledge recognition results. Ground-truth labels were established through a machine-assisted pre-labeling and human-verification procedure: the system first generated preliminary labels, which human annotators then validated against item information and item ontology definitions to produce final gold labels. The resulting samples are aggregated into the \textbf{item knowledge test set}, constituting the unified evaluation benchmark for subsequent model iterations.

Following this evaluation protocol, {\ssd} achieves 92\% precision and 78.3\% recall. In terms of item knowledge asset gain, compared with merchant-provided data, the average number of attributes per SKU increased to 1.5$\times$ the original level, while the total volume of item-knowledge assets (SKU $\times$ attribute key $\times$ attribute value) expanded by a factor of 3.35, reaching hundreds of billions. Of this total, merchant data accounts for 30\% and AI-generated data for 70\%. These results demonstrate that {\ssd} effectively enhances item knowledge coverage while maintaining high recognition accuracy.

Although the methods in Sections~\ref{sec:3_2} and~\ref{sec:4_2} have addressed item knowledge discovery and recognition, stable and controllable iteration remains an open challenge. In Section~\ref{sec:5}, we consolidate and extend these capabilities into the item understanding LLMs/VLMs with a controllable, continuously evolving training framework.

\subsection{Throughput Efficiency Improvement}

The {\ssd} architecture effectively resolves the scalability challenges associated with item knowledge recognition. Nevertheless, in industrial-scale production environments, tens of billions of items still impose stringent requirements on data freshness and system throughput. Consequently, throughput optimization must be grounded in a rigorous analysis of the intrinsic characteristics of large-scale e-commerce data. We summarize three key observations:

\begin{itemize}
    \item \textbf{Homogeneity of item features.} A substantial number of redundant SKU listings exist for identical physical items (e.g., the iPhone 17 Pro Max).
    
    \item \textbf{Skewed attribute distribution.} Common attributes are invoked frequently, whereas long-tail attributes, such as ``cuff pleat type'', are extremely sparse.
    
    \item \textbf{Repetitive computation across SKUs.} Multiple SKUs associated with the same SPU share the vast majority of core knowledge, resulting in redundant recognition computations if processed in isolation.
\end{itemize}

These observations inform our primary optimization objective: shifting the computational paradigm from ``all SKUs $\times$ all attributes'' to ``differentiated SKUs $\times$ highly relevant attributes''. This transformation substantially reduces compute costs while preserving recognition accuracy. To this end, we enhance throughput through three complementary strategies: computational load reduction, cache reuse, and asynchronous pipeline parallelism.

\subsubsection{Computational Load Reduction}

Computational load reduction implements this paradigm shift across two dimensions: SKU-level load reduction and attribute-level load reduction.

\paragraph{(1) SKU-level load reduction.}
SKU-level load reduction condenses ``all SKUs'' into a smaller set of differentiated SKUs. In large-scale item catalogs, many SKUs are homogeneous variants characterized by highly similar item information and attributes. To reduce computational redundancy, the system first determines through semantic retrieval whether the current item can reuse existing attribute recognition results.

Specifically, given an item $x$, we use the encoder $f_{\text{enc}}$ trained in Section~\ref{sec:4_2_1} to derive its semantic representation. We then construct a category-stratified vector retrieval index. Rather than conducting an exhaustive search over the entire item space $S$, the system constrains retrieval to the same-category candidate set $S_c \subset S$, where $|S_c| \ll |S|$. Within this reduced candidate set, the system computes cosine similarity to retrieve the Top-1 most similar item.

If the similarity score of the Top-1 item exceeds a predefined threshold, the system directly reuses that item's existing attribute recognition results for the current item. This mechanism effectively reduces redundant computations caused by homogeneous SKU variants.

\paragraph{(2) Attribute-level load reduction.}
Attribute-level load reduction further recasts ``full-attribute scanning'' as ``highly relevant attribute probing''. Rather than running {\ssd} across all attributes within a category, the system adaptively constrains the attribute space through two relevance-probing tasks: single-SKU attribute relevance probing (Task A) and SKU-pair attribute relevance probing (Task B).

\subparagraph{Task definition and data construction.}
To enable attribute-level load reduction, we train an attribute relevance-probing model prior to the {\ssd} stage. The training data are uniformly structured as discrimination triples $\langle x, K, K^* \rangle$, where $K$ denotes the complete set of attribute keys under a category and $K^*$ denotes the subset of highly relevant attribute keys predicted by the model. By injecting the relevance instruction $I_{\text{dis-k}}$, the task is formulated as:
\[
f_{\text{dis-k}}(x, K, I_{\text{dis-k}}) \rightarrow K^*, 
\quad \text{where} \quad K^* \subseteq K.
\]

The model jointly learns two sparsification strategies under a unified multi-task supervised fine-tuning framework:

\begin{itemize}
    \item \textbf{Task A: single-SKU attribute relevance probing.} 
    Given an item $p$ with item information $x_p$ and the complete category-level attribute key set $K$, the objective is to identify the subset of attributes $K^*(p)$ that are relevant to the item.
    
    \item \textbf{Task B: SKU-pair attribute relevance probing.} 
    Given a new item $p$ and a reference item $p'$ associated with the same SPU, the input $x$ is formulated as the differential fields $\Delta(x_p, x_{p'})$. The objective is to identify the subset of attributes $K^*_{\Delta}$ that are affected by the differences between $p$ and $p'$.
\end{itemize}

\subparagraph{Multi-task training and unified optimization.}
We integrate Task A and Task B into a unified multi-task SFT framework and use a dynamic mixing strategy to balance their data distributions during training. The model is optimized using the following autoregressive objective:
\[
\mathcal{L}
=
-\sum_{t}
\log P\left(
K^*_t
\mid
K^*_{<t}, x, K, I_{\text{dis-k}}
\right).
\]
This training objective encourages the model to acquire a generalized mapping from item semantics to attribute relevance, thereby enhancing its ability to determine which attributes require further processing by the {\ssd} pipeline.

\subparagraph{Inference.}
During inference, different capabilities are activated through task-specific instruction prefixes, enabling a unified model interface with task-adaptive behavior.

\begin{itemize}
    \item \textbf{Task A inference.} 
    For an individual item $p$, the model receives $x_p$ and the category-level attribute key set $K$ as input and generates the highly relevant attribute subset $K^*(p)$. The subsequent {\ssd} process is initiated exclusively for attributes within $K^*(p)$, whereas irrelevant attributes are assigned null values.
    
    \item \textbf{Task B inference.} 
    For an item pair $(p, p')$, where the attribute recognition results of the reference item $p'$ are accessible, the model uses the differential fields $\Delta(x_p, x_{p'})$ and the attribute key set $K$ as input to predict the affected attribute subset $K^*_{\Delta}$. Attributes excluded from this affected subset ($K \setminus K^*_{\Delta}$) directly inherit the recognition results of the reference item $p'$, thereby substantially reducing computational redundancy.
\end{itemize}

Through SKU-level semantic deduplication and attribute-level sparsification alone, the \oxygen production pipeline achieves a threefold improvement in throughput efficiency.

\subsubsection{Extreme Cache Reuse}
The {\ssd} inference prompt comprises task instructions, item information, attribute keys, and candidate values. In practice, different SKUs under the same SPU share approximately 85\% of their item information, such as item descriptions and other detailed technical content. Nevertheless, conventional approaches compute the entire prompt independently for each SKU, resulting in substantial redundancy in both memory consumption and computation. To mitigate this inefficiency, we implement SPU-level prefix cache reuse.

\begin{itemize}
  \item \textbf{Prompt structure optimization.} We place invariant information shared within an SPU, such as item descriptions, at the start of the prompt to establish a shared prefix, thereby maximizing prefix-cache hit rates.

  \item \textbf{Cache-aware locality guarantee.} Requests are grouped by SPU and routed to designated NPU devices, preventing frequent cache eviction caused by stateless load balancing.

  \item \textbf{Memory management tuning.} Given the structured nature of our prompts, we conduct source-code-level tuning of the \texttt{block\_size} parameter in vLLM \citep{kwon2023efficient}. By keeping cache hit rates high while limiting block-management overhead, we empirically identify \texttt{block\_size=16} as the optimal configuration for this scenario, compared with the default value of 128. This adjustment substantially enhances memory efficiency.
\end{itemize}

These optimizations alone improve production throughput by more than sixfold.

\subsubsection{Asynchronous Pipeline Parallelism}
The three-stage \oxygen production workflow, comprising vector generation, semantic search, and discrimination, requires coordination between heterogeneous hardware, specifically NPUs and CPUs. In this architecture, vector generation and discrimination are offloaded to NPUs, while semantic search is processed by a Faiss-based CPU cluster. However, data dependencies and inter-processor communication introduce idle intervals, resulting in compute resource underutilization and the formation of compute bubbles. To address these inefficiencies, we propose a hierarchical architecture that integrates L0 fine-grained data parallelism with an L1 cross-chunk asynchronous pipeline.

\begin{itemize}
  \item \textbf{L0 fine-grained data parallelism.} Rather than employing conventional coarse-grained SPU-level chunk partitioning, we partition large SPUs into fine-grained SKU chunks and implement dynamic load balancing. This prevents a single large chunk from occupying CPU resources for an extended period and blocking the entire pipeline.

  \item \textbf{L1 cross-chunk asynchronous pipeline.} We decouple sequential dependencies across chunks temporally, enabling the three stages to operate seamlessly and asynchronously. In the vector generation stage on NPUs, vLLM workers perform dynamic sharding and process different chunks according to available memory, eliminating the need to wait for an entire chunk to be ready. In the retrieval stage on CPUs, embeddings from multiple chunks are aggregated asynchronously to construct optimized batches, thereby maximizing vector database throughput. In the discrimination stage on NPUs, recalled sub-tasks are dynamically scheduled according to memory pressure, guaranteeing sustained utilization of NPU compute resources.
\end{itemize}

This architecture alone reduces waiting latency across heterogeneous hardware and achieves more than a twofold improvement in overall system throughput.

\subsubsection{Overall Effectiveness}
While each of the aforementioned strategies yields significant throughput gains, they are difficult to stack efficiently in practice. To assess the comprehensive benefits of the integrated system, we conducted controlled experiments under a uniform physical environment (a single Huawei Ascend 910C NPU), using the identical ontology and sampled item set. We measure throughput by the number of SKU $\times$ attribute key pairs processed per unit time per unit of compute, thereby reflecting the system's capability to perform large-scale knowledge inference. Under identical evaluation conditions, we executed the end-to-end inference pipeline before and after optimization and compared the total execution times to determine the throughput improvement ratio. All efficiency gains reported in this paper are computed according to this evaluation protocol.

By jointly applying the three-stage optimization strategies (computational load reduction, cache reuse, and asynchronous pipeline parallelism), the overall throughput efficiency is improved by more than tenfold. As the dynamically evolving ontology expands and the attribute dimension increases, all three optimizations exhibit strong scaling effects: a larger attribute space provides greater potential for sparsification and enhances prefix cache reuse. Consequently, system-level throughput gains are expected to become even more pronounced as the attribute dimension continues to expand.

\section{Item Understanding LLMs/VLMs}  \label{sec:5}

\begin{figure*}[t]
  \centering
  \includegraphics[width=0.8\textwidth]{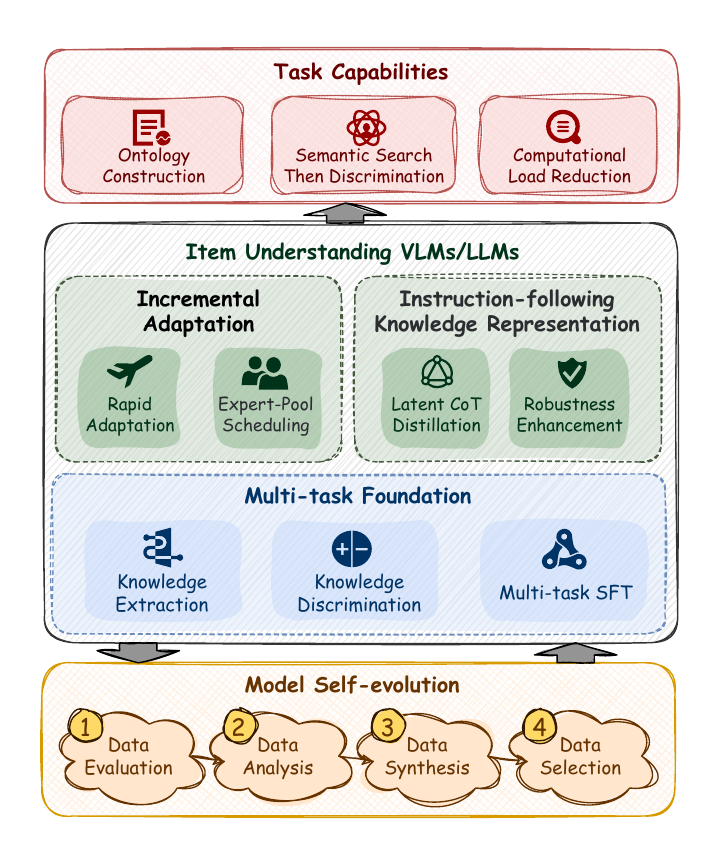}
  \caption{Overview of the \oxygen framework for the item understanding LLMs/VLMs. Constructed upon a unified multi-task item understanding foundation model, the framework supports incremental capability expansion, incorporates instruction-following knowledge representation, and implements a closed-loop model self-evolution mechanism to continuously enhance model performance and data quality.}
  \label{fig:item-understanding-large-model}
\end{figure*}

Along two dimensions---ontology engineering and the AI Item Library---the previous sections present an end-to-end pipeline that spans from dynamically evolving ontology construction to large-scale item knowledge production. Sustaining the long-term reliability of this system, while enabling its continuous evolution, requires a robust and controllable iteration framework. Building such a framework poses the following challenges:
\begin{itemize}
  \item Weak off-the-shelf performance in knowledge-intensive domains: \oxygen draws on extensive industry knowledge that off-the-shelf models lack, which limits their effectiveness~\citep{gururangan2020dont}.
  \item Costly full fine-tuning under frequent incremental updates: \oxygen continually expands with new domains, ontology entries, and tasks; static models cannot keep pace, while frequent full retraining is inefficient and risks catastrophic forgetting~\citep{mccloskey1989catastrophic,kirkpatrick2017overcoming}.
  \item Key features buried in strong noise: items carry rich information, and their implicit fine-grained features are easily obscured by irrelevant signals; this problem is especially pronounced for representation models.
  \item Subtle, hard-to-fix recognition defects: as the overall quality of item knowledge improves, the model still performs poorly on a small number of ontology recognition cases; such failures are subtle, rarely surface in aggregate metrics, cannot be resolved simply by adding training data, and instead require tailored data for repair.
\end{itemize}

To address these challenges, and building on the stable, controllable production pipeline established before, this section introduces the item understanding LLMs/VLMs framework of \oxygen (Figure~\ref{fig:item-understanding-large-model}). Developed upon a unified multi-task item understanding LLMs/VLMs foundation, it supports incremental capability expansion, instruction-following knowledge representation, and a closed-loop self-evolution mechanism, continuously enhancing model performance while supporting both ontology engineering and the AI Item Library.

\subsection{Multi-task Item Understanding LLMs/VLMs Foundation}
\oxygen originally trained separate models for scenarios such as knowledge discovery, semantic representation, and discrimination-based knowledge recognition. Although their training data all describe similar item--ontology semantics, such isolated training limited data reuse and knowledge transfer across capabilities while increasing model-management overhead \citep{ruder2017overview,raffel2020exploring}.

In the unified foundation, we organize the generative (non-representation) knowledge-production tasks across these scenarios into two families: knowledge extraction and knowledge recognition. We therefore reorganize the existing training data into a multi-task supervised fine-tuning (multi-task SFT) format \citep{wei2022finetuned,chung2024scaling} and consolidate these capabilities into the item understanding LLMs/VLMs:

\begin{enumerate}
  \item Knowledge extraction: autonomously identifies and extracts knowledge from item information, encompassing tasks such as attribute key extraction, attribute value extraction, and key-value extraction.
  \item Knowledge recognition: executes authenticity judgment or subset selection based on item information and a specified candidate space, including key discrimination, value discrimination, and key-value discrimination.
\end{enumerate}
Through unified modeling of these tasks, the item understanding LLMs/VLMs acquire a generalized understanding of items, ontology entries, and other task patterns, thereby serving as the foundation for \oxygen's model capabilities.

\subsection{Incremental Adaptation}
The multi-task foundation provides strong, general item understanding, but it exhibits limited generalization to continuously emerging ontology entries (such as a newly added ``pet food'' category), and frequent full retraining significantly reduces iteration efficiency.

\begin{figure*}[h]
  \centering
  \includegraphics[width=0.9\textwidth,height=0.52\textheight,keepaspectratio]{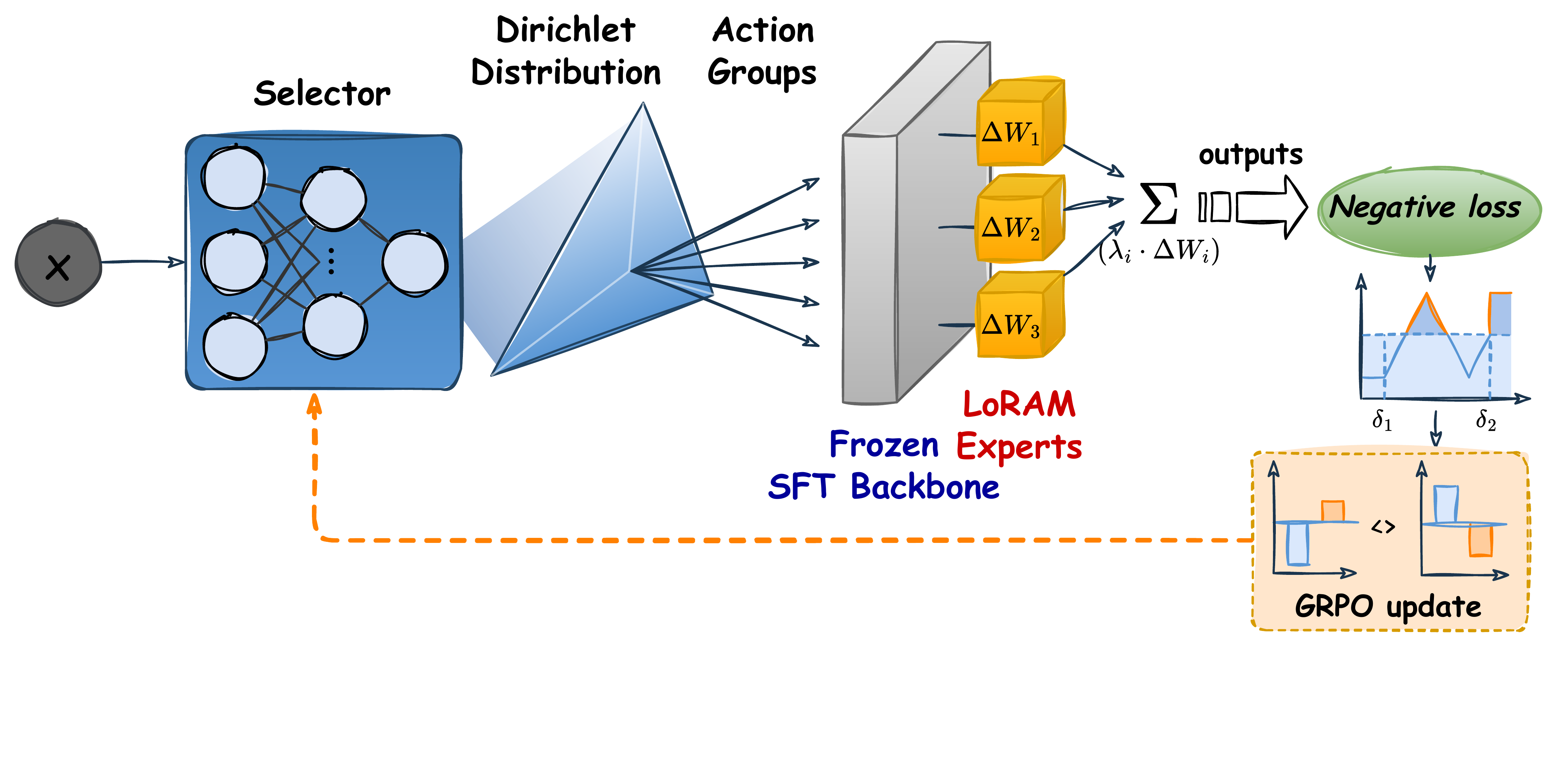}
  \caption{Incremental adaptation based on LoRAM experts and adaptive expert composition. A frozen SFT backbone is combined with multiple lightweight expert updates, and GRPO optimizes expert composition via task feedback.}
  \label{fig:incremental-adaptation}
\end{figure*}

To address this, we introduce a task-free, lightweight incremental learning mechanism to expand the model's capability boundaries without fully retraining the foundation model, while preserving existing capabilities \citep{houlsby2019parameter,biesialska2020continual}. The core method is to build on the robust multi-task foundation, develop lightweight ``expert modules'' for incremental requirements, and dynamically integrate them into the expert pool, enabling agile capability expansion. Figure~\ref{fig:incremental-adaptation} illustrates the incremental adaptation mechanism.

\paragraph{(1) Rapid adaptation to incremental tasks.} To keep up with ontology iteration, the system incrementally trains the corresponding expert modules, such as ``pet food'' and ``beauty and skincare'' domain experts.

Low-Rank Adaptation (LoRA) has become the preferred method for incremental adaptation owing to its high parameter efficiency \citep{hu2022lora}. It implements weight updates via the product of low-rank matrices: $\Delta W = \frac{\rho}{r} BA$, where $B \in \mathbb{R}^{n \times r}$, $A \in \mathbb{R}^{r \times m}$, and $r$ and $\rho$ represent the rank and the scaling factor, respectively.

However, conventional LoRA is constrained by its inherent low-rank structure, and the magnitude of its weight updates is significantly lower than that of full fine-tuning. This impairs the early-stage fitting dynamics, slows convergence, and prevents the model from reaching the desired accuracy under limited business data, even when the learning rate is increased. Consequently, we introduce LoRAM initialization based on the Magnitude Principle \citep{zhang2026loram}. By directly constructing an initial state with stronger update momentum, LoRAM accelerates convergence and enhances accuracy. The strategy uses the discrete sine transform (DST) to construct deterministic orthogonal bases $\xi$ and integrates them with a pretrained-weight-based gain coefficient $\beta$ to define the initial states of the low-rank matrices:
\[
B^{(0)} = \beta \cdot \xi_n, \quad A^{(0)} = \beta \cdot \xi_m^\top.
\]
To guarantee that the model's initial behavior matches that of the original foundation model, the system simultaneously performs weight compensation:
\[
W \leftarrow W - \beta^2 \xi_n \xi_m^\top.
\]
LoRAM optimizes the adapter's magnitude dynamics without additional memory overhead or preprocessing cost, improving the fitting efficiency while raising the performance ceiling for incremental requirements.

\paragraph{(2) Expert-pool scheduling.} LoRAM produces single-scenario experts. However, during inference, the model achieves higher recognition precision by dynamically aggregating multiple expert capabilities and sharing knowledge across all categories.

To this end, we propose GROLE, an incremental learning strategy based on adaptive LoRA expert composition \citep{liao2026grole}. We establish a modular expert pool in which each $\Delta W_{\text{LoRAM}, i}$ represents an independent capability unit. An adaptive selector $g_\phi$ estimates the fusion weights $\boldsymbol{\alpha}$ of incremental experts from input $x$ and task instruction $I$, with weights constrained to be nonnegative and to sum to one. The model output is determined by the linear combination of the foundation and experts:
\[
h = \left( W_0 + \sum_{i=1}^n \alpha_i \Delta W_{\text{LoRAM}, i} \right)x.
\]
where $W_0$ represents the multi-task foundation model and $n$ is the number of experts. This mechanism enables the model to achieve both general semantic understanding and specific logical judgment within a unified architecture.

Given that expert weights lack explicit labels, we employ Group Relative Policy Optimization (GRPO) \citep{shao2024deepseekmath} to model the allocation process as feedback-driven policy learning. The system samples $G$ groups of expert weights $\{\boldsymbol{\alpha}_j\}_{j=1}^G$ from a Dirichlet distribution and calculates relative advantages $\{A_j\}_{j=1}^G$:
\[
A_j = \frac{r_j - \text{mean}(\boldsymbol{R})}{\text{std}(\boldsymbol{R})}.
\]
where $r_j$ denotes the overall reward of the $j$-th weight vector $\boldsymbol{\alpha}_j$ across mixed tasks (defined as the negative loss), and $\boldsymbol{R}$ is the set of rewards for the current group $\{r_j\}_{j=1}^G$. The selector identifies the optimal policy by optimizing:
\[
\mathcal{J}_{GRPO}(\boldsymbol{\phi}) = \frac{1}{G}\sum\limits_{j=1}^G \min\left[\rho_j A_j, \operatorname{clip}(\rho_j, 1-\epsilon, 1+\epsilon)A_j \right].
\]
where $\rho_j$ denotes the importance sampling ratio of the selector with respect to $\boldsymbol{\alpha}_j$, and $\epsilon$ is the clipping coefficient. This mechanism enables the flexible composition of novel knowledge and task logic while preserving foundation capabilities, thereby supporting the ongoing expansion of item understanding.

This approach enables \oxygen to iterate rapidly. Meanwhile, to share item knowledge across all categories, the system periodically performs full fine-tuning, consolidating distributed incremental capabilities back into the foundation model. This supports the continuous integration of expert capabilities and the long-term evolution of the foundation model.

\subsection{Instruction-following Knowledge Representation}
In contrast to the generation-oriented item understanding LLMs/VLMs, knowledge representation aims to build a unified semantic space that maps item information, user queries, standardized ontology entries, and related information into the same vector space, supporting efficient alignment and retrieval at an industrial scale.

\begin{figure*}[h]
  \centering
  \includegraphics[width=1.0\textwidth,height=0.42\textheight,keepaspectratio]{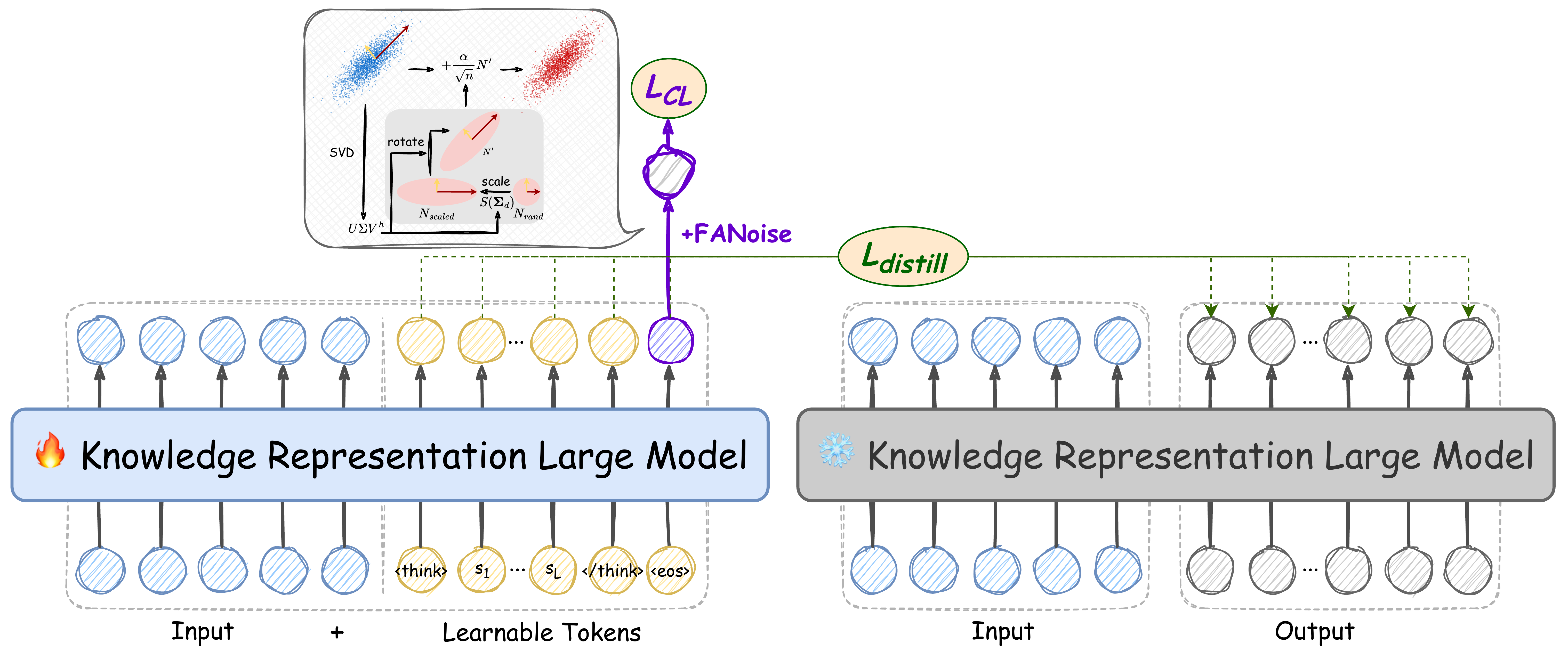}
  \caption{Instruction-following knowledge representation training. The framework transfers reasoning capability through latent chain-of-thought (Latent CoT) distillation and enhances representational robustness via adaptive feature-space perturbation.}
  \label{fig:knowledge-representation-training}
\end{figure*}

The knowledge representation component illustrated in Figure~\ref{fig:knowledge-representation-training} is trained to follow task instructions while maintaining robust fine-grained semantic alignment.

In e-commerce, representation models must extract knowledge signals from comprehensive item information. Traditional embeddings are susceptible to interference from literal similarity and noise, often failing to capture fine-grained details. For example, a product detail page may obscure the ``waterproof level'' within text such as ``waterproofing has been further upgraded, supporting IP68-grade dust and water resistance, and passing the T\"UV S\"UD 2-meter, 24-hour test." To address this challenge, we propose a ``Reasoning-then-Embedding'' framework that performs reasoning prior to generating an embedding vector, distinguishing essential knowledge from redundant noise \citep{wei2022chain}.

The framework is underpinned by two mechanisms. First, a reasoning-based representation mechanism built on Latent CoT distillation \citep{deng2023implicitcot} transfers the teacher model's reasoning capability into the hidden states of the student model. Second, a spectral-structure-based adaptive noise injection strategy \citep{li2026fanoise,guo2026spectral} dynamically adjusts perturbation intensity according to the representation distribution, strengthening the model's semantic robustness. Collectively, these mechanisms enable complex semantic modeling while remaining both efficient and accurate.

\paragraph{(1) Implicit reasoning representation.} In retrieval scenarios, representations need instruction-following capability to suppress irrelevant information and noise. To avoid the high latency associated with explicit CoT decoding \citep{wei2022chain}, following InstEmb \citep{gao2026instemb}, we adopt implicit reasoning, completing the reasoning process within a single forward pass.

We append $L$ learnable tokens $s=[\langle\text{think}\rangle, s_1, s_2, \dots, s_L, \langle/\text{think}\rangle, \langle\text{eos}\rangle]$ after the input sequence $x$ as carriers for reasoning. During training, a frozen teacher model provides supervision \citep{hinton2015distilling}. The inputs are defined as follows:
\[
x_{\text{student}} = [x; s], \quad x_{\text{teacher}} = [x; r_{cot}]
\]
where $r_{\text{cot}}$ is the explicit reasoning chain generated by the teacher model. For example, for the attribute ``waterproof level'', the teacher model can infer from wording such as ``...supports IP68-grade dust and water resistance...'' that ``IP68-grade'' is the core information, guiding the student model to internalize deeper semantics.

We design a position-aligned distillation loss that constrains the student model's hidden state $h^S_i$ at implicit-token positions to align with the teacher model's hidden state $h^T_{i-1}$ when processing the corresponding reasoning logic:
\[
L_{\text{distill}}= \frac{1}{L}
\sum_{i=|x|+2}^{|x|+L+1}
\left|
h^S_i- 
h^T_{i-1}
\right|_2^2.
\]
During inference, the hidden state of the last token $\langle\text{eos}\rangle$ in the implicit sequence is extracted as the reasoning-based representation vector: $e = h^S_{\langle\text{eos}\rangle}$. Without introducing autoregressive overhead, this method compresses core knowledge from unstructured descriptions into vector representations and enhances recall accuracy for fine-grained knowledge.

\paragraph{(2) Representation robustness enhancement.} After the reasoning-based representation $e$ is obtained, high-frequency marketing terms and redundant descriptions in e-commerce text may still cause the model to overfit superficial noise, weakening generalization in practical alignment scenarios. To address this, we propose a spectral-structure-based adaptive noise injection strategy at the feature layer to enhance representation robustness.

Given the batch representation set $E \in \mathbb{R}^{B \times d}$, we perform singular value decomposition (SVD): $E = P\Sigma Q^\top$. The singular value matrix $\Sigma$ reflects the energy distribution across semantic directions. Based on this structure, we dynamically adjust noise intensity: perturbation is increased in dominant directions, which often correspond to high-frequency noise patterns, to suppress overfitting, whereas perturbation is constrained in weak-signal directions to preserve fine-grained logical signals. The noise modulation process is defined as follows:
\[
N_{\text{scaled}} = N_{\text{rand}} \odot S(\Sigma_{\mathrm{diag}}), \quad \text{where } S \in \left\{ \frac{\Sigma_{\mathrm{diag}}}{\overline{\Sigma_{\mathrm{diag}}}}, \frac{\sqrt{\Sigma_{\mathrm{diag}}}}{\sqrt{\overline{\Sigma_{\mathrm{diag}}}}} \right\}.
\]
where $N_{\text{rand}}$ denotes Gaussian noise, $\Sigma_{\mathrm{diag}}$ is the diagonal singular-value vector, and $\overline{\Sigma_{\mathrm{diag}}}$ represents the mean singular value. The noise is subsequently projected back to the representation space through a basis transformation and normalized by dimension to generate the perturbed reasoning representation $\tilde{e}$:
\[
N' = N_{\text{scaled}} Q^\top,\quad \tilde{e} = e + \frac{\delta}{\sqrt{n}} N'.
\]
where $\delta$ is the global noise intensity coefficient and $n$ is the feature dimension.

In the contrastive learning stage, we follow the InfoNCE loss defined in Section~\ref{sec:3_2_2} and replace $e$ with $\tilde{e}$ to explicitly strengthen the semantic alignment:
\[
L_{\widetilde{\text{InfoNCE}}} = -\frac{1}{|V^+|} \sum_{i=1}^{|V^+|} \log \frac{\exp(\operatorname{sim}(\tilde{e}, \tilde{e}_i^+)/\tau)}{\exp(\operatorname{sim}(\tilde{e}, \tilde{e}_i^+)/\tau) + \sum_{j=1}^{|V^-|} \exp(\operatorname{sim}(\tilde{e}, \tilde{e}_j^-)/\tau)}.
\]
The final training objective comprises the distillation loss and the modified InfoNCE loss, where $\zeta$ is a weighting coefficient:
\[
L_{\text{total}} = L_{\text{distill}} + \zeta \cdot L_{\widetilde{\text{InfoNCE}}}.
\]
In summary, the model accurately extracts core semantics from complex e-commerce item information and achieves high-quality alignment and retrieval of atomic knowledge in vector space.

\subsection{Model Self-evolution}
As \oxygen developed, the item understanding LLMs/VLMs have achieved relatively stable performance on tasks such as knowledge extraction and recognition. However, in practical business scenarios, model issues gradually shift from explicit errors to more subtle long-tail defects. Relying solely on full-data expansion or periodic fine-tuning makes stable, low-cost continuous optimization difficult.

We consequently develop a self-evolution framework for the item understanding LLMs/VLMs \citep{zelikman2022star,madaan2023selfrefine}. The process comprises four stages (data evaluation, data analysis, data synthesis, and data selection) and establishes a closed-loop system through model iteration.

\begin{figure*}[h]
  \centering
  \includegraphics[width=1.0\textwidth,height=0.38\textheight,keepaspectratio]{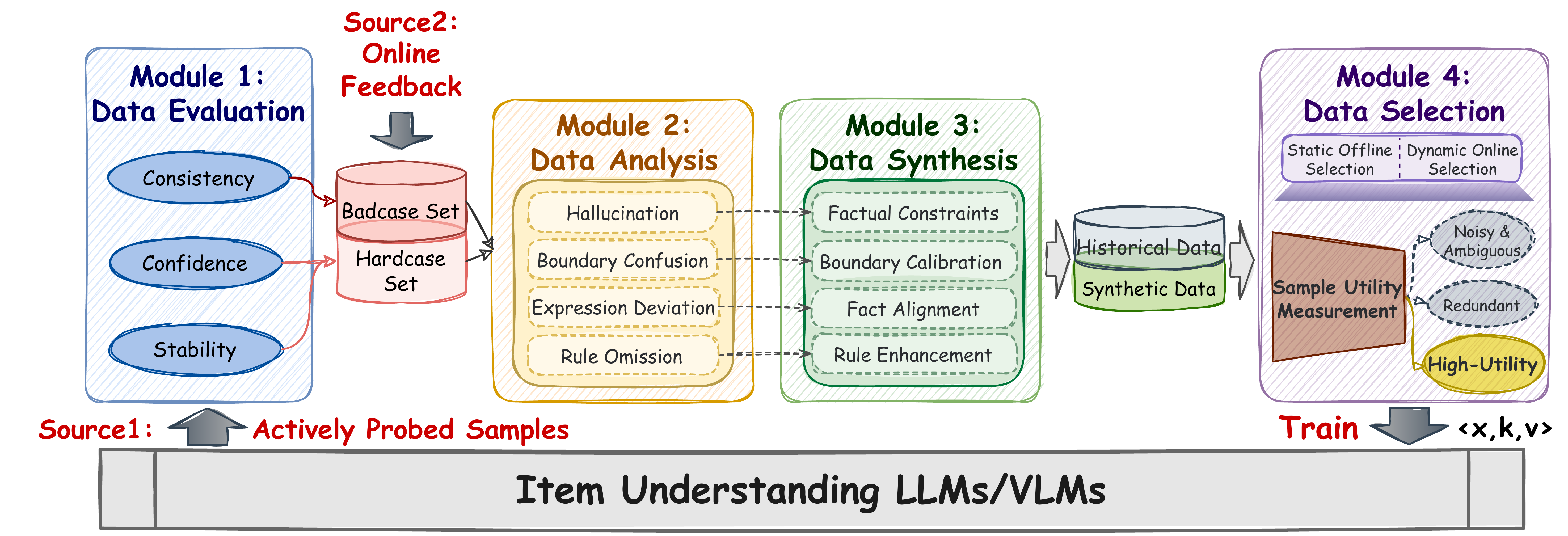}
  \caption{Model self-evolution framework. Online feedback and proactive mining yield bad cases and hard cases, which are analyzed, converted into targeted synthetic data, prioritized based on sample value, and used for controlled model iteration.}
  \label{fig:model-self-evolution}
\end{figure*}

Figure~\ref{fig:model-self-evolution} provides a system-level view of the self-evolution loop; the following paragraphs delineate its four modules.

\paragraph{Module 1: Data evaluation.} The data evaluation module identifies bad cases and hard cases from online feedback and actively probed samples. Samples are assessed along three dimensions: evidence consistency, model confidence, and perturbation stability.

First, the system implements prompt-constrained evidence consistency judgment, namely an LLM-as-a-judge mechanism \citep{zheng2023judging}, to validate the triples $\langle x, k, v \rangle$ produced by the model. Here, $x$ denotes item information, $k$ denotes the attribute key, and $v$ denotes the attribute value. The evaluation criterion is whether $v$ can be rigorously derived from the current input $x$ alone. If the evidence is contradictory or unsupported, the sample is assigned to the bad-case set.

Second, the system performs confidence estimation based on the model output distribution under greedy decoding. At each decoding step \(i\), the system computes token-level confidence as the probability margin between the top-ranked token and the second-ranked candidate token across the vocabulary $V$, and subsequently determines sequence-level confidence:
\[
C_i = \max_{j \in V} P_i(j) - \underset{k \in V}{\text{secondmax}} P_i(k),
\]
\[
C_{\text{avg}} = \frac{1}{N} \sum_{i=1}^N C_i.
\]
where $N$ is the output sequence length. Low overall confidence indicates ambiguity among multiple candidate answers, and the sample is designated as a hard-case candidate.

Finally, the system checks stability under input perturbation. While preserving core semantics, it introduces minor perturbations to the order of candidate attributes and item descriptions. If the output varies, the sample is also designated as a hard-case candidate.

The data evaluation module yields two sample types: bad cases and hard cases. Bad cases are used for defect attribution and corrective data synthesis, while hard cases help identify the model's capability boundaries.

\paragraph{Module 2: Data analysis.} The data analysis module uses large models to perform defect attribution for bad cases and hard cases. For bad cases, it analyzes erroneous outputs to identify underlying defect causes. For hard cases, it investigates sources of instability and categorizes samples for corrective training, boundary enhancement, or manual review.

Common defects include evidence-missing hallucination, where the input lacks explicit evidence for the target attribute while the model predicts an attribute value based on category commonsense; attribute-boundary confusion, where the model identifies valid evidence but incorrectly associates it with a semantically similar attribute, resulting in an incorrect key-value binding; attribute-value expression deviation, where the output drifts semantically from the original evidence or over-refines it by introducing unexpressed modifiers; and missing category-specific rules, where the same attribute has different judgment criteria across categories and the model fails to capture category-context differences.

Using ``lampshade material'' as an example, if the item information only describes a ``Nordic-style desk lamp'' but the model outputs ``wood'', this is evidence-missing hallucination. If the text contains ``metal lamp base, glass lampshade'' but the model associates ``metal'' with ``lampshade material'', this is attribute-boundary confusion. If the source text only states ``glass material'' but the model outputs ``frosted glass'', this is attribute-value expression deviation.

The module outputs the defect type, impact scope, and suggested repair strategy, transitioning model iteration from indiscriminate data expansion to defect-targeted remediation.

\paragraph{Module 3: Data synthesis.} The data synthesis module uses large models to synthesize targeted training samples based on identified defect types and repair suggestions. Rather than merely increasing data volume, its primary goal is to produce information-dense repair samples.

For evidence-missing hallucination, the system employs constrained synthesis to generate input samples lacking explicit evidence for the target attribute while constraining the target output to ``not mentioned'' or ``cannot determine'', thereby reinforcing the model's awareness of evidence boundaries. For attribute-boundary confusion, the system uses boundary-calibration synthesis to formulate contrastive samples targeting easily confused attributes, helping the model learn the correct bindings between attributes and evidence spans. For attribute-value expression deviation, the system implements fact-alignment synthesis to generate strictly consistent standardized outputs based on factual anchors in the source input, preventing the introduction of extraneous semantics during extraction. For missing category-specific rules, the system applies category-rule enhancement synthesis, integrating category ontology definitions, attribute value spaces, and historical error distributions to generate training samples that align with the business interpretations of specific categories.

For example, for the ``lampshade material'' issue, the system generates negative samples devoid of material evidence and trains the model to output ``not mentioned''. It also constructs contrastive samples containing both ``metal lamp base'' and ``glass lampshade'' to help the model distinguish the attribute boundaries between diverse components.

To ensure the quality of synthetic data, all samples undergo evidence consistency validation and format validation; high-risk samples are further evaluated via manual spot checks. The module generates candidate training data accompanied by defect labels, repair objectives, and sample-source markers.
 
\paragraph{Module 4: Data selection.} The data selection module identifies high-utility samples from synthetic samples, historical samples, and online feedback samples to enhance iteration efficiency and model performance.

We introduce a sample utility measurement strategy grounded in the cognitive gap \citep{xie2023doremi,wang2026tandem,wang2026blade}. For a candidate sample $x$, we calculate the losses of the current model $u$ and the expert model $w$ on the sample, and define the sample utility score as:
\[
L_{\text{excess}}(x) = \mathcal{L}_{\mathbf{u}}(x) - \mathcal{L}_{\mathbf{w}}(x).
\]
where the current model $u$ represents the active online model or the model to be optimized, and the expert model $w$ can be a higher-capacity teacher model, a strong baseline model, or a domain model.

A high $L_{\text{excess}}$ indicates that the current model has not yet mastered the sample, whereas the expert model exhibits proficiency, making the sample highly informative and valuable for training. If both losses are low, the sample is likely redundant and may be reserved for distribution preservation or deferred. If both losses are high, the sample may suffer from label noise, insufficient evidence, or semantic ambiguity, and is subsequently directed to the hard-case pool for manual verification.

Data selection operates in two modes. Static offline selection evaluates the candidate sample pool at scale prior to training and constructs a specialized training set. Dynamic online selection adaptively adjusts training batches based on model convergence during incremental training, allowing the model to prioritize the hardest and most informative samples.

The module categorizes candidate samples based on their respective roles in the subsequent iteration:
(1) \textbf{Training set.} High-utility samples are incorporated into the training set for model optimization.
(2) \textbf{Current-round test set.} Representative bad cases and stability hard cases identified in the current iteration are maintained as test samples for the attributes currently being optimized. After each iteration, these samples are annotated via the procedure described in Section~\ref{sec:4_2_3} and then merged into the item knowledge test set, ensuring the benchmark continuously expands and supports all subsequent evaluation scenarios. Samples identified as redundant, low-utility, ambiguous, insufficiently grounded, or confirmed noisy are excluded from both training and validation.

\subsection{Metrics}
Throughout the continuous iteration of the item understanding LLMs/VLMs, we conduct offline validation by computing metrics on end-to-end item knowledge production results using the unified \textbf{item knowledge test set}.

After incorporating multi-task SFT, incremental scenario adaptation, instruction-following knowledge representation training, and model self-evolution, the latest model attains 94.2\% precision and 82.8\% recall in end-to-end AI Item Library production. Compared with the results in Section~\ref{sec:4_2_3}, precision and recall exhibit increases of 2.2\% and 4.5\%, respectively.

The evaluation results further demonstrate that only 0.8\% of attributes undergo a precision degradation of more than 5\%, indicating that the model enhances overall item knowledge production quality while maintaining stable performance on the continuously growing benchmark.

For newly added ontology entries, the end-to-end turnaround time from attribute mining to finalized production data is reduced from over 30 days to approximately two weeks, representing a substantial improvement in production efficiency.

\subsection{Platform Capabilities}
During the large-scale deployment of \oxygen, the underlying compute platform encounters two primary technical challenges: model training and inference on Huawei Ascend NPUs, and the efficient use of compute resources. At the model level, the compute platform offers a unified training and inference framework compatible with Huawei Ascend NPUs, ensuring the robust and efficient execution of the end-to-end \oxygen pipeline. At the resource level, the compute platform incorporates elastic resource scheduling for offline NPU clusters, enabling compute resource reuse across the production workflow and improving overall compute utilization.

\section{Item Tunnel}

To support ontology construction, AI Item Library production, and model self-evolution, while enabling efficient consumption by downstream applications, we designed and implemented \oxygen's unified item tunnel as a data and compute hub between production and applications.

The system must address four engineering challenges. First, there is an inherent trade-off between data freshness and cost, while downstream consumers differ substantially in freshness requirements, item scope, and ontology coverage. 
Second, the evolution of the AI Item Library relies on multiple hybrid stream-batch pipelines; long cascades across heterogeneous engines may introduce distributed-state synchronization problems and data inconsistencies.
Third, data flows covering tens of billions of items consume CPU, NPU, and distributed-storage resources across the full stack, so even local inefficiencies can affect global stability and compute-cost amortization.
Fourth, business scenarios share a common foundation, take multimodal inputs, and differ only in service details; ad hoc customization would fragment interfaces, undermine contract governance, waste resources, and weaken SLA guarantees.

To address these challenges, we developed the item tunnel, a unified infrastructure for large-scale knowledge management. The item tunnel supports hundreds of millions of AI Item Library updates per day, delivers tiered freshness spanning seconds, minutes, and days, and maintains eventual consistency across heterogeneous service modalities with convergence achieved within minutes. Powered by this infrastructure, \oxygen has built a dynamic AI Item Library covering tens of billions of items. The system is now deployed across a broad spectrum of production applications, including search, advertising, marketing, shopping assistants, item operations, product listing, and platform governance.

The following four engineering practices correspond to the four challenges above.

\begin{itemize}

\item \textbf{Tiered-freshness pipelines: inference tasks are assigned based on data-change frequency and business value to provide freshness tiers from seconds to days.} Inference is the key factor governing both freshness and cost: offline inference has high throughput and NPU utilization but longer end-to-end latency, whereas real-time inference responds quickly but may repeatedly process transient intermediate states and amplify compute cost \citep{kwon2023pagedattention,agrawal2024sarathi}. We therefore divide computation into three paths. The offline path uses batch computation and offline inference to build a low-cost item-knowledge baseline, and decouples CPU processing and NPU inference into a pipeline to improve NPU utilization. The nearline path refreshes minute-level increments through micro-batches. The real-time path is reserved for high-value, highly time-sensitive items and triggers online inference only after staged pruning, retrieval, and discrimination. Together, the real-time, nearline, and offline paths deliver second-, minute-, and day-level freshness, respectively, balancing latency, freshness, and cost \citep{akidau2015dataflow}.

\item \textbf{Eventual consistency: the system resolves data-version divergence across cascaded production paths and controls convergence time by freshness tier, down to the minute level.} The tiered-freshness design is essentially a hybrid stream-batch, Lambda-style architecture \citep{marz2015bigdata,akidau2015dataflow}, so it must prevent incorrect version overwrites and data loss caused by distributed production, replay, or delayed data. At the same time, it must guarantee convergence under the eventual-consistency model used in large distributed stores \citep{vogels2009eventually,decandia2007dynamo}. We achieve consistency through three mechanisms. For merge-conflict resolution, we introduce monotonically increasing version markers: streaming results use event time and batch results use snapshot time. Based on Hudi's MVCC and ACID capabilities, the system applies last-write-wins semantics at the item-attribute granularity, so higher versions overwrite lower versions \citep{apachehudi2026concurrency,apachehudi2026recordmerger,aws2026globaltables}. For computational consistency, operators are separated into stateless and stateful classes; stateful operators always read the latest item state, and Spark partition-level atomic commit, together with Flink checkpoint persistence and alignment, provides At-Least-Once recovery from persisted checkpoints \citep{zaharia2012rdd,carbone2017state}. For concurrency control, compute and storage partitions are aligned by item ID, and read-write locks plus scheduler mutual exclusion prevent read-write conflicts and duplicate processing. High-freshness scenarios therefore converge within minutes, while the remaining data is exposed through daily fully consistent snapshots.

\item \textbf{Storage-compute efficiency: physical decoupling and elastic scheduling support stable production over data flows covering tens of billions of items.} Heterogeneous cross-cluster coordination between Spark/Flink and the vLLM inference cluster depends on three mechanisms. First, storage and computation are aligned via a unified data contract: offline production is anchored on Parquet-formatted shards in HDFS, real-time production is anchored on Kafka partitions, and upstream computation pre-aggregates by category and SPU to produce strict-schema shards \citep{apacheparquet2026overview,kreps2011kafka}. Second, vLLM workers dynamically claim work shards and reuse the prefix cache, turning fixed assignment into a dynamic pipeline and reducing the impact of straggler nodes \citep{kwon2023pagedattention}. Third, checkpointed operator state is keyed by item ID, so after a failure the system retries only unfinished shards at fine granularity. This design preserves compute efficiency while preventing data loss. Overall, this yields a production system that provides on-demand resource allocation, retryable tasks, and end-to-end monitoring.

\item \textbf{Unified services: a standard service matrix replaces fragmented integrations and supports scalable reuse across business scenarios.} On top of algorithmic and engineering infrastructure, the tunnel unifies offline storage, containerization, inference engines, and service frameworks, and exposes two standard service classes. Item data services provide online lookup of ontology and item instances, large-scale relational tables for high-throughput consumption, and event-driven full or incremental change streams, covering both static retrieval and dynamic subscription. Item algorithm services use the item understanding LLMs/VLMs to embed algorithmic capabilities into item-lifecycle stages such as product listing validation and compliance governance, and expose high-precision category prediction, fine-grained attribute recognition, and long- and short-title generation as online AI services. This turns scenario-specific model use into platform-level reuse.

\end{itemize}

\section{Applications}

\oxygen has constructed millions of ontology entries and hundreds of billions of high-quality item-knowledge assets, covering all major JD categories. To make these core knowledge assets broadly usable across diverse business scenarios, \oxygen standardizes and productizes its foundational resources through the item tunnel's unified delivery layer, exposing them as a portfolio of reusable capabilities. This design closes the loop across scenarios, roles, and feedback channels, making \oxygen a strategic digital knowledge infrastructure for high-quality business growth across the group.

\subsection{Consumer-facing Applications}
By improving the quality and richness of item-information at the source, \oxygen has been integrated into core consumer shopping flows. It supports more precise traffic distribution, improves the shopping experience, and better serves users' diverse and personalized needs as well as emerging forms of intelligent shopping interaction.

\textbf{Search.} \oxygen has been integrated into core search traffic-allocation stages, including recall, relevance ranking, and query understanding. It improves the quality of item information and raises its richness to 3.35$\times$ the previous level, reducing the share of item-information defects in search by 37\% and thereby lowering the overall search bad-case rate. The AI Item Library and ontology also support search guidance and faceted filtering, helping users find desired items more quickly and accurately.

\textbf{Item detail pages.} \oxygen improves the content displayed on product detail pages by extracting the core structured information that users care about and presenting it in a form that is easy to read and understand. Traditional detail pages often contain inconsistent information across touchpoints, repeated content, and overly technical specifications, all of which increase users' decision-making cost. \oxygen mitigates information conflicts at the source, surfaces intelligent short titles, core selling points, and core attributes, and adds AI explanations, such as product-function summaries and parameter comparisons. These capabilities turn dense technical specifications into plain-language descriptions, helping users quickly grasp an item's core value and improving both user experience and conversion efficiency.

\textbf{AI shopping assistants.} With \oxygen as the item-knowledge foundation, AI shopping assistants can access structured and standardized knowledge across categories, enabling more accurate understanding of both items and user intent, driving a smarter shopping experience. Representative applications include:
\begin{itemize}
  \item \textbf{Conversational e-commerce.} Moving beyond keyword search and static shopping pages, conversational e-commerce uses large-scale structured item knowledge to handle fragmented, scenario-based, and long-tail shopping needs through natural-language dialogue. Users can find items, check specifications, and ask about matching or usage scenarios through casual conversation.
  \item \textbf{AI comparison.} The platform automatically identifies users' comparison needs and initiates an AI-powered comparison flow. It generates structured reports covering product highlights, core parameters, and user reviews, helping users compare candidate items efficiently and make decisions with less effort.
\end{itemize}

\subsection{Merchant and Business Operations}
For item management across JD's domestic and international businesses, self-operated and platform sellers, and B2C/B2B/O2O formats, \oxygen combines the AI Item Library with AI-based item understanding and generation capabilities to form an end-to-end operational loop spanning category planning, product listing, and product operations. This loop improves operational efficiency, raises data quality, and supports more precise traffic acquisition.

\textbf{Category planning.} Through intelligent analysis of platform-wide data, \oxygen shortens the decision cycle from two or three weeks to a few days and helps identify category growth opportunities more precisely. Built on the standardized item-knowledge system in the AI Item Library, and combined with user behavior, industry trends, and on- and off-site market-performance data, this capability is delivered to merchants and category buyers as a productized tool. Integrated AI explanations and automated reports surface supply-demand gaps, category competition, and user-demand preferences, helping operators identify growth opportunities and design differentiated operating strategies.

\textbf{Product listing.} The automated fill rate of core attributes exceeds 80\%, improving item-data quality at the source while reducing operating costs for category buyers and merchants. With the item understanding LLMs/VLMs as the model foundation and the AI Item Library and ontology as the knowledge foundation, \oxygen enables AI-assisted product listing with end-to-end information pre-filling. Merchants and category buyers only need to upload a main image or a title, and the system automatically performs category recognition, brand recognition, and attribute filling. 

\textbf{Product operations.} Based on category expertise, \oxygen optimizes item creatives at scale and increases click-through rate by about 9\%. Through multimodal learning, the system distills optimal visual standards, including resolution, color, composition, and information hierarchy, from high-conversion item images. It then standardizes and enhances image creatives, reorganizes shopping-guide information, and iterates through A/B testing, achieving click-through gains at very low compute cost. Copy optimization reuses the same standards to accumulate industry best practices and support merchant self-service optimization. In addition, a fully managed AI service allows category buyers and merchants to delegate bulk creative optimization.

\subsection{Platform Operations}
For platform-side scenarios involving ecosystem governance and platform-wide resource orchestration, \oxygen drives a shift from experience-driven operations to data-driven and AI-assisted decision-making. It covers core scenarios such as merchant recruitment and assortment review, campaign page construction, audience operations, targeted advertising, product information governance, and price governance. By linking supply and demand, \oxygen increases the utilization and value of item assets across the platform, improves operational efficiency and marketplace governance, and supports the long-term healthy growth of the platform ecosystem.

\textbf{Marketing operations.}
\begin{itemize}
  \item \textbf{Assortment selection.} The assortment selection platform improves product information quality and richness across all categories, supporting more efficient marketing operations. Traditional assortment selection is often constrained by poor product information quality and fragmented information systems, leading to incomplete or inaccurate selection results. The AI Item Library enriches product information across categories and is integrated with the assortment selection platform. During rule configuration, it recommends categories and attributes to avoid manual omissions, reduce configuration cost, and improve both the richness and precision of assortment results.
  \item \textbf{Audience profiling.} By combining product information with user behavior, \oxygen supports fine-grained audience profiling and attribute-based identification of high-potential users, improving advertising conversion and lowering customer acquisition cost. Item attributes accumulated in the AI Item Library are linked with user-behavior data to characterize segmented customer groups. These attributes can also support reverse audience discovery, producing precise audience segments, such as high-potential interest groups and competitor-intent groups, for on-site and off-site advertising, push notifications, and other targeted operations.
\end{itemize}

\textbf{Platform-ecosystem development.}
\begin{itemize}
  \item \textbf{Product information governance.} Through an end-to-end item-information governance loop, we substantially reduce the share of item-information-related negative exposure, improving platform-wide item-information compliance and quality. The system establishes full-lifecycle product information governance, proactively intercepts non-compliant submissions during product listing, and, after products are listed, uses intelligent inspection tools to automatically verify items across the platform, identify category and attribute errors in bulk, and prompt category buyers and merchants to correct them.
  \item \textbf{Price governance.} \oxygen provides identical-item recognition with over 90\% accuracy across all physical categories in JD Retail, helping maintain a fair and orderly pricing environment. Category buyers need to compare prices during sourcing, procurement, and selling-price decisions to keep prices competitive while protecting profit margins. At the same time, the platform uses a price-rating system to guide merchants and category buyers toward more competitive and standardized pricing. Built on high-dimensional product information and image-text information in the AI Item Library, the identical-item recognition capability supports price governance, identical-item price comparison, price control during major promotions, and abnormal-price detection.
\end{itemize}

\section{Related Work}
Research on e-commerce item knowledge falls into four lines: e-commerce knowledge graphs, ontology expansion, item knowledge production, and e-commerce domain foundation models. Existing works have made notable progress, but most focus on isolated point solutions rather than providing a top-level, systematic solution to the integrated infrastructure required for the production, quality inspection, consumption, and feedback of large-scale item knowledge. Without that closed-loop foundation, knowledge assets cannot be reused and improved efficiently across large-scale business scenarios.

\subsection{E-commerce Knowledge Graphs}
Early e-commerce knowledge graph research mainly focused on semantic relations among items, concepts, and user needs. AliCoCo pointed out that the traditional category--property--value (CPV) system is insufficient to express real shopping needs. Therefore, it introduced ``e-commerce concepts'' as intermediate semantic entries to connect high-level intents such as ``outdoor barbecue'' and ``gifts for the elderly'' with item organization \citep{luo2020alicoco}. AliCoCo2 further extended this to e-commerce commonsense relation modeling, characterizing commonsense mappings among scenarios, concepts, and item features so that graphs can better support search rewriting, recommendation expansion, and semantic matching \citep{luo2021alicoco2}.

Subsequently, research began to mine implicit intent knowledge from user behavior. FolkScope combines the generative capabilities of LLMs with a human-in-the-loop process to distill purchase intents from co-purchase behavior and build an intent knowledge graph \citep{yu2023folkscope}. COSMO scales e-commerce commonsense knowledge production to industrial settings through a pipeline of ``LLM generation--human-feedback-trained critic/classifier--small-model scalable generation'' \citep{yu2024cosmo}. In addition, item knowledge graphs have been applied to item relation modeling, recommendation, and explainable ranking. Representative studies include systematic graph construction pipelines with taxonomy enrichment, knowledge extraction, and quality control \citep{zalmout2021all}, as well as relation learning for complements, substitutes, and co-viewed items based on user behavior and multimodal product information \citep{xu2020pkg_embedding}. Recently, LLM-PKG further explored distilling recommendation relations and explanations inferred by LLMs into item knowledge graphs to reduce the risk of hallucination in generative models \citep{wang2024llmpkg}.

Overall, these works improve the structured representation of item relations, user intents, and e-commerce commonsense. However, they focus on individual stages of graph construction or application, leaving closed-loop coordination among knowledge production, validation, deployment, and feedback insufficiently explored.

\subsection{Ontology Expansion}
Ontology and category taxonomy expansion form another important direction in e-commerce knowledge construction. Since e-commerce catalogs involve rapidly changing item types, complex category hierarchies, and dynamically evolving attribute systems, statically and manually maintained taxonomies are difficult to sustain for large-scale item understanding. General taxonomy expansion research has proposed various automated methods, such as taxonomy expansion based on hierarchical topic models \citep{zhang2018taxogen}, HiExpan's task-guided taxonomy construction \citep{shen2018hiexpan}, TaxoExpan's framework for predicting parent-child nodes when inserting new concepts into an existing taxonomy \citep{shen2020taxoexpan}, and STEAM's mini-path-based self-supervised taxonomy expansion \citep{yu2020steam}.

In e-commerce scenarios, Octet formulates taxonomy expansion as a taxonomy enrichment problem for online item catalogs and uses heterogeneous relations among queries, items, and categories for self-supervised training to adapt to continuously emerging new item types \citep{mao2020octet}. KATIE further focuses on category-attribute relation discovery, attribute importance modeling, and attribute synonym merging, showing that e-commerce ontology expansion has shifted from simple entry expansion to fine-grained schema learning for search, recommendation, and item understanding \citep{errahmadi2023katie}.

Overall, related studies mainly address how to expand an ontology and which attributes are applicable to a given category. However, these methods model ontology evolution, attribute schema recognition, and SKU-level knowledge completion separately, making it difficult to achieve efficient system-level coordination.

\subsection{Item Knowledge Production}
One core task of item knowledge production is to automatically extract attribute values from titles, descriptions, parameter tables, and images. OpenTag formulates attribute value extraction as an open-world extraction problem, eliminating the dependence on closed-set dictionaries \citep{zheng2018opentag}. SUOpenTag extends extraction tasks to thousands of attributes for industrial-scale scenarios \citep{xu2019scaling}. AVEQA formalizes attribute value extraction as a question-answering task, treating attributes as questions and extracting answer spans from item contexts, thereby improving adaptability to large-scale and unseen attributes \citep{wang2020aveqa}.

As item detail pages contain increasingly rich sources of information, research has further shifted toward multi-source and multimodal attribute extraction. MAVE constructs a large-scale multi-source item attribute value extraction dataset \citep{yang2022mave}. Subsequent studies jointly model text and image signals for attribute prediction and attribute value extraction \citep{zhu2020multimodal}, and further explore cross-category visual attribute extraction \citep{lin2021pam}, structured multimodal Transformer encoding \citep{wang2022smartave}, and large-scale multimodal attribute extraction based on generative question answering \citep{khandelwal2023mxt}. These works show that item knowledge completion has gradually evolved from text sequence labeling into a unified extraction problem involving open attributes, multi-source inputs, and multimodal fusion.

Beyond extraction, industrial-scale item knowledge systems must also determine whether extraction results are trustworthy. Existing work formulates attribute value verification as a low-resource task to identify whether attribute values are consistent with item descriptions \citep{wang2020automatic}. In recent years, LLMs have also been used for attribute extraction, normalization, and catalog quality governance. Representative studies include evaluations of GPT-series models for item attribute extraction and normalization \citep{brinkmann2024llm_pave}, CatalogRAG for retrieval-augmented LLM-based attribute completion \citep{zhang2025catalograg}, combining brand knowledge bases with LLM agents for attribute repair and item matching \citep{ceker2026brand_agent}, and multimodal self-correcting instruction tuning for open attribute discovery \citep{li2025msit}.

Overall, attribute extraction and knowledge completion have progressed from open extraction and large-scale label expansion to multi-source modeling, multimodal fusion, and automatic validation. However, most methods devote limited attention to production and management efficiency at massive scale, which limits their feasibility in industrial deployments.

\subsection{E-commerce Domain Foundation Models}
As large language models and large multimodal models have advanced, researchers have begun to build e-commerce domain foundation models to reduce fragmentation across task-specific models. LLaMA-E introduces instruction tuning into e-commerce scenarios, covering tasks such as ad generation, title rewriting, item classification, purchase intent inference, and question answering \citep{shi2025llamae}. eCeLLM further constructs the ECInstruct dataset, covering attribute extraction, item relation prediction, item matching, sentiment analysis, sequential recommendation, query-item ranking, item question answering, and other tasks, demonstrating the generalization capability of instruction-tuned LLMs in e-commerce scenarios \citep{peng2024ecellm}.

At the same time, research has also begun to introduce e-commerce knowledge during pretraining. LiLiuM adapts the tokenizer, training data, and multilingual capabilities for eBay's e-commerce setting \citep{herold2024lilium}. e-Llama adapts a general foundation model to the e-commerce domain through continued pretraining \citep{herold2025ellama}. Compass-v3 trains an MoE domain model for multilingual e-commerce scenarios in Southeast Asia, reflecting the development trend toward multilingual, domain-specific, and scalable deployment \citep{maria2025compassv3}.

For multimodal item understanding, MOON proposes a generative multimodal representation learning framework for e-commerce to handle multiple images paired with a single text description, background noise, and multi-task transfer \citep{zhang2026moon}. MOON2.0 further improves item multimodal representation learning through dynamic modality balancing, multi-granularity image-text alignment, and collaborative image-text enhancement \citep{nie2026moon20}. These works show that e-commerce domain models are moving from single-task fine-tuning toward multi-task, multilingual, multimodal, and scalable foundation models.

Although e-commerce domain foundation models have significantly improved item semantic understanding, cross-task transfer, and cold-start generalization, they still focus primarily on model capabilities. In contrast, industrial-scale item knowledge infrastructure must also address dynamic ontology evolution, knowledge production costs, result verifiability, hybrid online-offline serving, and downstream feedback loops. Therefore, integrating domain foundation models, ontology, and the AI Item Library into a sustainably evolving item knowledge system remains a key challenge for both research and industrial deployment.

\section{Conclusion}
This paper presents the JD Oxygen AI Item Center (\oxygen), an industrial-scale, LLM/VLM-centric infrastructure for item knowledge production and serving in large-scale e-commerce scenarios. \oxygen establishes an end-to-end system spanning ontology construction, item knowledge production, model evolution, item tunnel, and downstream business applications.

For ontology engineering, \oxygen combines expert domain knowledge with the generalization and reasoning capabilities of large models through efficient human--AI collaboration. This enables the dynamic discovery, fusion, validation, and continuous expansion of the ontology, forming a high-quality, comprehensive, and timely updated item knowledge backbone. 

For the AI Item Library, \oxygen adopts a ``semantic search then discrimination'' architecture, decoupling the dynamic ontology from model parameters so that newly added ontology entries can be quickly incorporated into the production pipeline. With computational load reduction, cache reuse, and asynchronous pipeline parallelism, \oxygen achieves high-throughput and low-cost item knowledge production at the scale of tens of billions of SKUs.

For the model system, \oxygen builds a multi-task large model for item understanding and combines incremental learning, instruction-following representations, and model self-evolution. This allows the model to steadily improve in a constantly changing item ecosystem, repair long-tail defects, and avoid systemic degradation. 

On the engineering side, the item tunnel delivers AI-generated item knowledge efficiently, securely, and reliably to a wide range of business scenarios, including search, recommendation, product listing, governance, operations, merchant recruitment and assortment review, and identical-item recognition, through tiered freshness pipelines, eventual consistency guarantees, and a unified service framework.

\oxygen currently supports knowledge production across tens of thousands of categories and tens of billions of SKUs at JD, accumulating hundreds of billions of item-knowledge assets and delivering significant business impact in item management, traffic distribution, and platform operations. Our practice shows that, for industrial-scale e-commerce knowledge construction, relying solely on isolated model capabilities is insufficient for scalable deployment. Only by systematically coordinating large-model capabilities, ontology engineering, knowledge production, engineering systems, and business feedback loops can we build a scalable and continuously evolving item knowledge infrastructure. The development of \oxygen validates the industrial feasibility of LLMs/VLMs for large-scale e-commerce item understanding and provides a reusable technical path for future AI-driven item management, intelligent operations, and e-commerce infrastructure upgrades.

\section{Limitations and Future Work}
Although \oxygen V1 has delivered significant results at its current stage of development, ultra-large-scale industrial deployment still poses several fundamental challenges. (1) Ontology engineering: the current version has substantially expanded the ontology's conceptual coverage. However, relation modeling within the ontology remains preliminary and has not yet reached scale. This limits \oxygen's potential to empower applications and makes it difficult to store and use JD's accumulated industry knowledge efficiently. (2) AI Item Library: although the generated knowledge is already of high quality, the scale of tens of billions of items inevitably leads to numerous online failure cases that affect user experience. Making these defects detectable and correctable in a timely manner therefore remains the central challenge. (3) Item understanding LLMs/VLMs: supervised fine-tuning can lead to catastrophic forgetting. The key challenge is to improve domain-specific capabilities without degrading the model's general capabilities.

In future work, we will address these challenges along three directions. First, we will strengthen ontology engineering by expanding relation modeling, enriching the ontology, and enabling graph-based reasoning. We will further incorporate expert e-commerce knowledge and industry know-how into \oxygen, so that domain experience can be reused efficiently across business scenarios. Second, we will develop an online mechanism for bad-case discovery. Coupled with a data flywheel and model self-evolution, this mechanism will allow knowledge consumption to feed back into and drive knowledge production. Finally, we will investigate more effective ways to mitigate catastrophic forgetting during domain adaptation. Potential directions include scaling the number of experts in MoE architectures, simplifying task complexity through improved problem formulation, and enhancing prompts with explicit knowledge injection.

\bibliographystyle{ACM-Reference-Format}
\bibliography{main_en_v2_review_refs}
\end{document}